\documentclass[format=acmsmall]{acmart}
\usepackage{booktabs} % For formal tables
\usepackage[utf8]{inputenc}
\usepackage{adjustbox}
\usepackage{paralist}
\usepackage{graphicx}
\usepackage{subcaption}
\usepackage{multirow}
\usepackage{bbm}
\usepackage{amsmath}
\usepackage{amsfonts}
\usepackage{url}
\usepackage{color}
\usepackage{mathtools}
\usepackage{wrapfig}
\usepackage[ruled,linesnumbered,lined,boxed]{algorithm2e}
\usepackage{array}
\usepackage{bm}
\usepackage{acronym}
\usepackage{enumerate}
\usepackage[shortlabels]{enumitem}
\usepackage[colorinlistoftodos]{todonotes}
\usepackage{arydshln}

\AtBeginDocument{%
  \providecommand\BibTeX{{%
    \normalfont B\kern-0.5em{\scshape i\kern-0.25em b}\kern-0.8em\TeX}}}

\acrodef{GBS}{Generalized Binary Search}
\newcommand{\RQ}[2]{
    \begin{description}[topsep=0pt,nosep=0pt, leftmargin=0.75cm, noitemsep,nolistsep]
    \phantomsection\label{section:setup:rq#1}
    \item[RQ#1] #2
    \end{description}
}
\newcommand{\RQRef}[1]{\textbf{\hyperref[section:setup:rq#1]{RQ#1}}}
\newcommand{\ie}{\emph{i.e.,}\xspace}
\newcommand{\eg}{\emph{e.g.,}\xspace}

\let\cal\mathcal
\acmJournal{TOIS}
\acmYear{2022}
\setcopyright{acmlicensed}
\acmDOI{XXX/XXX.XXX}

\begin{document}

\title{Learning to Ask: Conversational Product Search via Representation Learning}

\author{Jie Zou}
\affiliation{%
  \institution{University of Amsterdam}
  \city{Science Park 904, 1098 XH Amsterdam}
  \country{The Netherlands}}
\email{j.zou@uva.nl}
\author{Jimmy Xiangji Huang}
% \authornote{Corresponding author}
\affiliation{%
  \institution{York University}
  \city{4700 Keele St, Toronto}
  \country{Canada}}
\email{jhuang@yorku.ca}
\author{Zhaochun Ren}
% \authornotemark[1]
\affiliation{%
  \institution{Shandong University}
  \city{72 Binhai Highway, Qingdao, Shandong Province}
  \country{China}}
\email{zhaochun.ren@sdu.edu.cn}
\author{Evangelos Kanoulas}
\affiliation{%
%   \institution{ILPS}
  \institution{University of Amsterdam}
  \city{Science Park 904, 1098 XH Amsterdam} 
  \country{The Netherlands}
%   \postcode{01003}
}
\email{e.kanoulas@uva.nl}
\authorsaddresses{Jie Zou, University of Amsterdam, Science Park 904, 1098 XH Amsterdam, The Netherlands, j.zou@uva.nl; Jimmy Xiangji Huang (corresponding author), York University, 4700 Keele St, Toronto, Canada, jhuang@yorku.ca; Zhaochun Ren (corresponding author), Shandong University, 72 Binhai Highway, Qingdao, Shandong Province, China, zhaochun.ren@sdu.edu.cn; Evangelos Kanoulas, University of Amsterdam, Science Park 904, 1098 XH Amsterdam, The Netherlands, e.kanoulas@uva.nl.}

\renewcommand\shortauthors{Zou, J. et al}



\begin{abstract}
Online shopping platforms, such as Amazon and AliExpress, are increasingly prevalent in the society, helping customers purchase products conveniently. With recent progress on natural language processing, researchers and practitioners shift their focus from traditional product search to conversational product search. Conversational product search enables user-machine conversations and through them collects explicit user feedback that allows to actively clarify the users' product preferences. 
Therefore, prospective research on an intelligent shopping assistant via conversations is indispensable. Existing publications on conversational product search either model conversations independently from users, queries, and products or lead to a vocabulary mismatch. 
In this work, we propose a new conversational product search model, ConvPS, to assist users to locate desirable items. The model is first trained to jointly learn the semantic representations of user, query, item, and conversation via a unified generative framework. 
After learning these representations, they are integrated to retrieve the target items in the latent semantic space. 
Meanwhile, we propose a set of greedy and explore-exploit strategies to learn to ask the user a sequence of high-performance questions for conversations. Our proposed ConvPS model can naturally integrate the representation learning of the user, query, item, and conversation into a unified generative framework, which provides a promising avenue for constructing accurate and robust conversational product search systems that are flexible and adaptive. Experimental results demonstrate that our ConvPS model significantly outperforms state-of-the-art baselines.
\end{abstract}


\maketitle

\section{Introduction}
The rapid growth of digital markets and the widespread use of e-commerce platforms have led to a surge of research activity in the field of e-shopping~\cite{rowley2000product,zhang2018towards,zou2019learning}. A key functionality of e‐shopping platforms is product search, which aims to support users to effectively locate products that they wish to buy~\cite{rowley2000product}. 

Traditional product search systems usually require customers to formulate queries and browse through the resulted products to locate the target items (\ie items they finally purchase). Searching and browsing turn out to be inefficient. Often this is due to the fact that the customers' descriptions of products do not match the vendors' descriptions ~\cite{li2014semantic}. 
The experience may become worse for mobile users since scanning a long list of products is impractical on limited bandwidth interfaces~\cite{zamani2020generating}. This leads to an increasing concern as mobile users continue to rapidly expand around the world. Fortunately, the rise of intelligent assistants (\eg Google Now, Apple Siri, and Microsoft Cortana) and dialogue systems provide new interaction modes between humans and systems through conversations. This lays the groundwork for a conversation-based intelligent assistant for product search, with the aim to alleviate the burden of reformulating queries and browsing through dozens of products, as well as offer a more natural way to dictate preferences and product characteristics. 
Compared to traditional product search, conversational product search systems interact with users by natural language and collect explicit feedback from users to clarify users' needs, leading to a better understanding of users’ dynamic preferences. 

Existing publications on product search mostly focus on traditional product search, using exact term matching methods~\cite{ponte1998language} or semantic matching functions~\cite{van2016learning,ai2017learning} to match the query with the product directly or in the latent space. 
Conversational product search, on the other hand, remains a challenging open problem.~\citet{zhang2018towards} presented the first conversational model for product search, in which a unified approach was proposed for conversational product search and recommendation by asking questions over aspect-value pairs extracted from user reviews. Similarly, ~\citet{bi2019conversational} also collected user feedback on the aspect-value pairs while they proposed a paradigm for conversational product search based on negative feedback. 
Instead of item aspects,~\cite{zou2019learning} asked questions based on extracted informative terms and proposed a question-based Bayesian product search model. Although these aforementioned approaches demonstrate their success to some extent, there are still some limitations. 
~\citet{bi2019conversational} focused on learning from negative feedback from shown items and model conversations \textit{fully independently}.~\cite{zou2019learning} updated user preferences and selected products on the basis of lexical overlap between terms in clarifying questions and product descriptions, leading to a \textit{hard-matching problem}, \ie ignoring the fact that a term may not appear in the description of a product but a synonym term or passage may make the product still relevant.~\citet{zhang2018towards} treated users, queries, products, and conversations in the same way by creating a concatenation of word embeddings to express user preferences, which \textit{neglects the hierarchical structure} of users, queries, products, and conversations, and may suffer from the \textit{vocabulary mismatch}. 

To relax the above limitations, in this work we propose a \textbf{Conv}ersational \textbf{P}roduct \textbf{S}earch model, ConvPS, to assist users in reaching their target items interactively. We carefully design our model as a unified generative model, which integrates representation learning of user, query, item, and conversation into a unified framework (rather than model them fully independently). To this end, the hard-matching problem, hierarchical structure problem, and vocabulary mismatch problem are also effectively alleviated by conducting product retrieval via soft-matching based on the jointly learned user, query, item, and conversation representations in the learned hierarchical latent semantic space. The probability of observed user-query-conversation-item quadruples can be deduced directly from their distributed representations, making the whole framework explainable and extendable. 
Our ConvPS model maintains a separate conversation representation and then combines it into the user representation and query representation to update the ranked list of items. Specifically, our ConvPS model (1) constructs a question pool based on slot-value pairs (aspect-value) following ~\cite{bi2019conversational, zhang2018towards}, (2) learns the representation of users, queries, items, and conversations in terms of slot-value pairs, (3) learns to ask a sequence of high-performance questions (slots) to the user based on a greedy or an explore-exploit strategy, (4) collects user feedback and enriches the user query with conversation representations, and (5) retrieves target items by using the user representation and the enriched query representation.
    
The major contribution of this paper is three-fold:
\begin{enumerate}
    \item First, we propose a new conversational product search model, ConvPS, that integrates the representation learning of user, query, item, and conversation into a unified generative framework, and thus retrieving the relevant items based on these jointly learned representations in the latent semantic space.  
    \item Second, we propose four greedy and explore-exploit learning to ask strategies to select a sequence of high-performance questions to ask, perform analysis to compare these learning to ask strategies and validate their effectiveness.
    \item Third, we propose a framework that can effectively incorporate two independent modules: the representation learning module for retrieving items and learning to ask module for selecting questions to ask.
\end{enumerate}
Our experiments on the product search datasets demonstrate that our ConvPS model significantly improves the performance of product search compared to state-of-the-art baselines. 

The rest of this paper is organized as follows. In Section~\ref{sec:relwork}, we discuss the related work. 
Section~\ref{sec:meth} introduces the details of our approach. Section~\ref{sec:exp} describes the experimental setup, research
questions, experimental results and corresponding analysis, while Section~\ref{sec:conc} concludes the paper. 

\section{Related Work}
\label{sec:relwork}
In this section, we summarize the related work which falls into three categories: product search, learning to ask, and conversational search. There is a large number of studies on the aforementioned topics. Here we only review work closely related to our research presented in this article. 

\subsection{Product Search}

Compared with ad-hoc retrieval tasks (\eg Web search) for finding relevant documents~\cite{DBLP:journals/isci/HeHZ11,DBLP:journals/ir/HuangPSCR03}, product search focuses on locating not only relevant products but also relevant products a user is willing to purchase~\cite{wu2018turning,ai2017learning}, making the task more challenging. Given a user query, multiple products could be relevant but only a few are actually purchased. Further, often product descriptions come with metadata (\eg brand name, types, and categories), leading to a semi-structured information space. Considerable work has been done on product search based on structured product information~\cite{lim2010multi}. Despite the effectiveness structured knowledge can offer, Vandic et al.~\cite{vandic2012faceted,vandic2013facet} found free-form user queries do not make use of structured knowledge available on product information pages. Instead, there is a vocabulary gap between product specifications and search queries~\cite{duan2013supporting, duan2013probabilistic}, with the concepts expressed by different terms in the product description and user queries~\cite{van2016learning, 10.1145/3269206.3271668}. To alleviate the problem of the vocabulary gap, plenty of attempts on representation learning have been introduced. For example, Duan et al.~\cite{duan2013supporting, duan2013probabilistic} proposed a probabilistic mixture model based on query generation via two language models.~\citet{van2016learning} presented a latent vector space model to learn representations of queries and products, so that they can be mapped to the same latent space to calculate the similarity between them. Then~\citet{ai2017learning} noticed the importance of personalization in product search and thus proposed a hierarchical embedding model by incorporating user representations for personalized product search. Later,~\citet{xiao2019dynamic} proposed a Dynamic Bayesian Metric Learning model to solve the problem of personalized product search under a streaming scenario.~\citet{ai2019zero} and~\citet{guo2019attentive} then applied an attention mechanism to improve the performance of personalized product search. On the other hand,~\cite{ai2019explainable} focused on the explainability and constructed an explainable retrieval model for product search. 
~\citet{ahuja2020language} proposed a multi-lingual multi-task learning framework and studied product search in a cross-lingual information retrieval context. The effectiveness of incorporating external information for product search is also explored by some work~\cite{guo2018multi,zhang2019neural,wang2020metasearch,bi2019study}. 
Besides representation learning techniques, there are a variety of studies on extracting different features and feeding them into learning-to-rank models~\cite{karmaker2017application, hu2018reinforcement}. Other factors such as diversity~\cite{yu2014latent} and user satisfaction~\cite{carmel2020people,lee2008effects} are also studied. 
Similar to previous publications, in this paper, we also attempt to alleviate the language gap between product-related documents (\eg product description and user reviews) and user queries based on representation learning. Different from the aforementioned publications that treat product search as a non-interactive problem, we propose a conversational product search approach, where ranking is conducted by a dynamic process, \ie the ranking in the next iteration incorporates the conversation based on the fine-grained feedback from users.
  
Recently, conversational product search has gained interest within the information retrieval community. ~\citet{zhang2018towards} treated conversational search and recommendation together and proposed a first conversational product search model with multi-memory network architecture. They ask questions over item aspects extracted from user reviews and collect user feedback on the corresponding values for the asked aspects. Similarly,~\citet{bi2019conversational} also constructed conversations and enquire user's explicit responses on the aspect-value pairs mined from product reviews while they highlight the value of negative feedback in conversational product search. Instead of item aspects,~\citet{zou2019learning} queried users on extracted informative terms (typically entities) from item-related descriptions and reviews, and proposed a sequential Bayesian method based on a cross-user duet training for conversational product search.~\citet{zou2020empirical} conducted an empirical study to quantify and validate user willingness and the extent of providing correct answers to the asked questions in question-based product search systems. In this work, we follow~\citet{zhang2018towards} and ~\citet{bi2019conversational} to construct conversations based on aspect-value pairs. Different from the above publications, we jointly learn the representations of users, queries, items, and conversations via a unified generative model. Moreover, we propose four greedy and explore-exploit learning to ask strategies to select a sequence of high-quality questions to ask to the user, and validate their effectiveness. 

\subsection{Learning to Ask}
The research on applications of learning to ask is broad. For example,~\citet{rao2018learning} learned to ask useful questions for clarification on community question answering.~\citet{hu2018playing} and~\citet{chen2018learning} investigated the optimal strategy of question selection on a 20 Questions game setup.~\citet{aliannejadi2019asking},~\citet{hashemi2020guided},~\citet{rosset2020leading} and~\citet{zou2020towardsa} focused on learning to ask informative and useful clarifying questions in information-seeking systems. ~\citet{christakopoulou2016towards},~\citet{ren2021learning} and~\citet{zou2020towards} aimed to select next questions to ask the user for conversational recommendation~\cite{10.1145/3477495.3531852}.~\citet{ruotsalo2018interactive} focused on selecting relevant keywords to display to the user for exploratory search. Although the importance of learning to ask in different tasks, learning to ask in product search is still relatively unexplored.~\citet{zou2019learning} learned to ask a good question based on user preferences and the rewards over question performances.~\citet{zhang2018towards} predicted the next question to ask to the user by maximizing the probability of the next question based on the softmax output layer for probability estimation. Compared with their greedy question selection strategies, we propose a set of systematic learning to ask strategies, including both greedy and explore-exploit strategies along with a comparative analysis among them. Further,~\citet{zou2019learning} asked questions about descriptive terms of items while we query users about aspects. ~\citet{zhang2018towards} asked questions over limited aspects, relying on the log-likelihood of probability estimation of the next aspect. Instead, we learn to ask questions over all aspects in the collection which is more challenging, based on one-hot representation in terms of aspect appearance in items.

Active learning~\cite{tong2001active} and bandit learning have also been used in the area of learning to ask. 
Active learning is mostly used for selecting informative items to form item-based questions to interact with users. For instance, \citet{iovine2021empirical} investigated the effects of several item selection strategies based on active learning in the conversational recommender scenario. \citet{christakopoulou2016towards} deployed several active learning strategies to identify the most appropriate items to ask the user for preference elicitation. 
They also conducted experiments to compare the performance of active learning based question selection techniques with bandit learning based approaches. As active learning methods query for labels that provide a high amount of new information, bandit learning approaches balance the need to explore new information with a focus on exploiting what has already been learned~\cite{auer2002using}. Such an explore-exploit balance may help focus on the most relevant questions, while still considering unexplored questions with great potential of high effectiveness. Following this line, \citet{christakopoulou2018learning} proposed a collaborative-bandit approach to tackle the explore-exploit trade-off in the problem of learning to interact with users. \citet{zhang2020conversational} and \citet{li2021seamlessly} proposed UCB methods or Thompson Sampling based model to deal with the explore-exploit trade-off in conversational recommender scenarios. In this paper, we propose both greedy and bandit learning strategies along with a comparison analysis among them in conversational product search.

\begin{figure}
  \includegraphics[width=1\columnwidth]{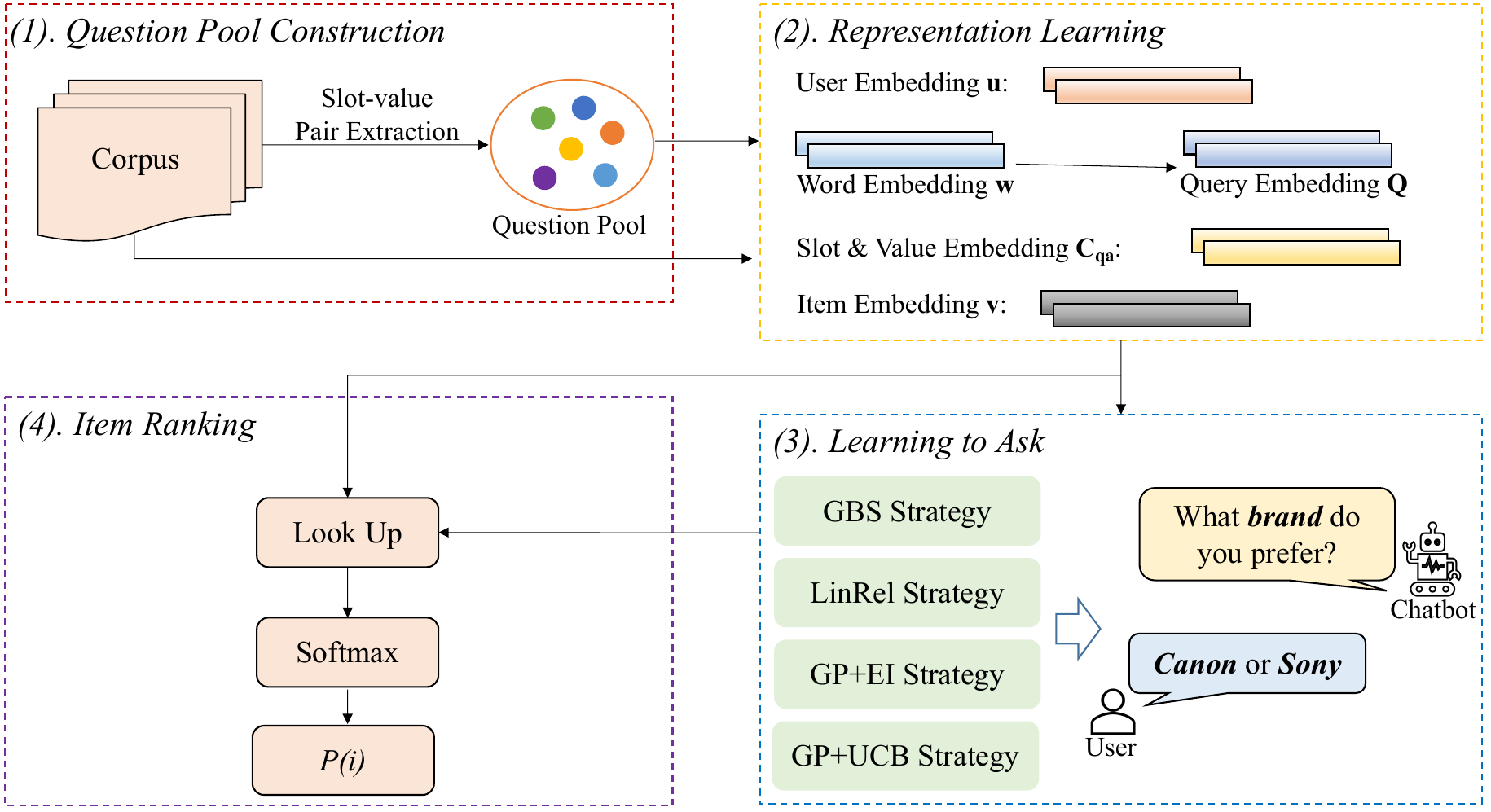}
  \caption{The research framework of our ConvPS model. (1) We first construct a question pool via slot-value pair extraction. (2) We then learn the user embeddings, conversation embeddings in terms of slot-value embeddings, item embeddings, and word embeddings to formulate query embeddings via our generative model. (3) Afterwards, we learn to ask a sequence of questions by the four proposed strategies: GBS strategy, LinRel strategy, GP+EI strategy, and GP+UCB strategy. (4) Lastly, we generate the item ranked list by estimating the probability of items based on these learned embeddings.}  
  \label{fig:framework}
\end{figure} 

\subsection{Conversational Search}
Early exploration of conversational search investigated mixed-initiative systems by interacting with users via script-based conversation during a search session~\cite{Belkin1995CasesSA}. Recently, researchers investigated conversational search by asking clarifying questions (CQs)~\cite{zamani2020generating,rosset2020leading,wang2021controlling}. ~\citet{radlinski2017theoretical} introduced a theoretical framework for conversational search and highlighted the importance of conversational search. Among previous studies about conversational search, they can be classified into four main categories: algorithms for conversational search~\cite{kenter2017attentive,zou2020towardsa,hashemi2020guided,zamani2020generating,wang2021controlling}, evaluation of conversational search~\cite{lipani2021doing,keyvan2022approach}, datasets for conversational search~\cite{qu2018analyzing,nguyen2016ms,dalton2020cast,aliannejadi2019asking,zamani2020mimics,aliannejadi2020convai3,ren2021wizard}, and empirical studies for conversational search~\cite{zamani2020analyzing,vtyurina2017exploring,kiesel2018toward,Trippas:2017:PIC:3020165.3022144,krasakis2020analysing,zou2020empirical,10.1145/3524110}. 
Compared with the aforementioned work on conversational search to retrieve documents or answers, our research focuses on product-seeking conversations.

\begin{table}[tb]
\caption{A running example. In this example user ``A1TR3G8YUR8QMG'' (represented by the user embedding $\mathbf{u}$) issued a query ``personal health foot care insert insole'' (represented by the query embedding $\mathbf{Q}$) and would like to locate the target item ``B003FJACI8'' (represented by the item embedding $\mathbf{v}$). Each item is annotated by a set of slot-value pairs, which is shown in the example indicated in blue. The corresponding embeddings of user, query, item, and conversation are indicated by square brackets. For each conversation round, we select the question with the optimal score calculated by our corresponding learning to ask function to ask the user. In our example, in the first round, we selected the question ``What \textcolor{blue}{price} (represented by the slot embedding $\mathbf{q}$) would you like?'', and received user response ``I want \textcolor{blue}{outstanding} (represented by the value embedding $\mathbf{a}$) price''. After the first round of conversation, the rank of target item increases from 2,251 to 522. }
\label{table:runningexample}
\begin{center}
\begin{tabular}{llr}
\toprule
\multicolumn{3}{p{0.7\columnwidth}}{\textbf{User:} }-- [$\mathbf{u}$]\\
\multicolumn{3}{p{0.7\columnwidth}}{ID: A1TR3G8YUR8QMG} \\
\hline
\multicolumn{3}{p{0.7\columnwidth}}{\textbf{Query:} ``personal health foot care insert insole'' } -- [$\mathbf{Q}$]\\
\hline
\multicolumn{3}{p{0.7\columnwidth}}{\textbf{Target item:}} -- [$\mathbf{v}$]\\
\multicolumn{3}{p{0.7\columnwidth}}{ID: B003FJACI8,} \\
\multicolumn{3}{p{0.7\columnwidth}}{Title: UGG Australia Men's Replacement Insoles Boot Inserts.}\\
\multicolumn{3}{p{0.7\columnwidth}}{Extracted slot-value pairs: (\textcolor{blue}{size}, \textcolor{blue}{big}), (\textcolor{blue}{foam}, \textcolor{blue}{thin}), (\textcolor{blue}{price}, \textcolor{blue}{outstanding}), (\textcolor{blue}{shipping}, \textcolor{blue}{free}), (\textcolor{blue}{quality}, \textcolor{blue}{real}), (\textcolor{blue}{insole}, \textcolor{blue}{thick}) ...
} \\
\hline 
\textbf{Conversation:} & & --[$\mathbf{c_{qa}}$]\\
System & User & Rank of target item\\
 & & 2251\\
 What \textcolor{blue}{price} [$\mathbf{q}$] would you like? & & \\
 & I want \textcolor{blue}{outstanding} [$\mathbf{a}$] price. & 522\\
 What \textcolor{blue}{quality} [$\mathbf{q}$] would you like? & & \\
 & I want \textcolor{blue}{real} [$\mathbf{a}$] quality. & 53\\
What \textcolor{blue}{brand} [$\mathbf{q^-}$] would you like? & & \\
 & No brand. & 34\\
 What \textcolor{blue}{size} [$\mathbf{q}$] would you like? & & \\
 & I want \textcolor{blue}{big} [$\mathbf{a}$] size. & 3\\
...\\
\bottomrule
\end{tabular}
\end{center}
\end{table}

\section{Methodology}
\label{sec:meth}
In this section, we introduce our proposed conversational product search model, ConvPS. A research
framework based on the ConvPS model is shown in Figure~\ref{fig:framework}. This framework includes four modules: (1) question pool construction; (2) representation learning; (3) learning to ask; and (4) item ranking. In the following, we will first describe the conversational product search problem and formally define our model, and then describe each of the four modules in detail. A running example is provided in Table~\ref{table:runningexample}. 

\subsection{Problem Formalization}
Assume we have $M$ users $\cal{U} = \{u_1, u_2, \dots, u_M\}$ and $N$ items $\cal{V} = \{v_1, v_2, \dots, v_N\}$. Each item $v_j \in \cal{V}$ is a product in e-commerce, and each item has its textual description $d_j$.  
A search session is initiated with a query $Q$ issued by a user $u_i$. During the search session, suppose we ask the user a clarifying question and the user provides an (associated) answer for the asked clarifying question, denoted as $(q', a')$. In the conversation for each search session, we ask the user a sequence of clarifying questions and collect a sequence of user answers for the asked clarifying questions. Suppose we asked $L$-round questions, the sequence of clarifying questions and answers are denoted as $\{(q'_1, a'_1) (q'_2, a'_2), \dots, (q'_L, a'_L)\}$. Thus, the sequence of actions in the conversation can be represented with: 
\begin{equation}
\label{qua:process}
u_i \to Q \to q'_1, a'_1, q'_2, a'_2, \dots, q'_L, a'_L, \to v^*_j,
\end{equation}
where $v^*_j$ is the target item the user is looking for. 
The goal of the system is, after collecting the $L$-iteration user’s feedback, to show a ranked list of items that ranks the target item $v^*_j$ on the top.

\begin{table}
\caption{A summary of key notations in this work.}
\label{tab:notation}
\centering
%  \small
\begin{tabular}{l|p{0.8\columnwidth}}
\toprule
 \textbf{Notation} & \textbf{Explanation}\\
 \hline
  $u_i$, $\cal{U}$ & The $i$-th user, and the set of all users, respectively \\
  $v_j$, $\cal{V}$ & The $j$-th item, and the set of all items, respectively \\
 $M, N$ & The number of users $M = |\cal{U}|$, and number of items $N = |\cal{V}|$ \\
 $Q$, $\mathbf{Q}$ & The initial user query and the embedding of a query, respectively\\
 $\mathbf{u}$  & The embedding of a user \\
 $\mathbf{v}$ & The embedding of an item \\
 $\mathbf{w}$ & The embedding of a word \\
 $(q, a)$ & The slot-value pair\\
 $\mathbf{c_{qa}}$ & The corresponding embedding of a slot-value pair $(q, a)$\\
 $\mathbf{x_q}$ & The one-hot vector of slot $q$ \\
  $q_l$, $L$ & The $l$-th selected slot and the length (\ie number of slots) of a conversation, respectively \\
  $\mathbf{q}$, $\mathbf{a}$ & The trained embeddings of a slot with positive feedback, and a value, respectively \\
  $\mathbf{q^-}$ & The trained embeddings of a slot with negative feedback \\
  $\cal{D}_w$ & The set of vocabulary of words in the corpus \\
  $\cal{D}_u$ & The words in textual descriptions of user $u$ \\
  $\cal{D}_v$ & The words in textual descriptions of item $v$ \\
  $\cal{S}_c$& The set of slot-value pairs in the corpus\\
  $\cal{S}_u$ & The slot-value pairs in the historical conversations of user $u$\\
  $\cal{S}_v$ & The slot-value pairs in the historical conversations of item $v$\\
  $\cal{S}_{uQv}$ & The slot-value pairs in the conversations of a user-query-item triple \\
\bottomrule
\end{tabular}
\end{table}
\subsection{Question Construction}
In this work, the questions in conversations are constructed based on clarification features, which are extracted or predefined from all the textual descriptions. This means we ask clarifying questions about clarification features by using templates to generate them, similar to existing publications ~\cite{bi2019conversational,zhang2018towards,zou2019learning,zou2020towards,zou2020towardsa,zamani2020generating}. 
To collect clarification features, in this work, we use slot-value pairs, and in particular a set of aspect-value pairs extracted from item descriptions and user reviews following the approach taken by~\citet{bi2019conversational} and~\citet{zhang2018towards}. Note that one could also use other features like keywords, item categories, or facets to construct questions. We leave the investigation of natural language generation (\ie generating natural language sentences from slot-value pairs) and natural language understanding (\ie understanding natural language sentences to extract slot-value pairs) as future work. Accordingly, the system can ask a sequence of questions in the form of ``what [slot/aspect] would you like?'', the users can respond with specific values corresponding to the asked slot, or ``not relevant'' as the negative feedback according to their information needs. We denote the extracted slot-value pair in the conversation as $(q, a)$. For example, (size, big) is a slot-value pair $(q, a)$ and it can construct a question ``what size would you like?'', as shown in our running example in Table~\ref{table:runningexample}.

\subsection{Representation Learning}
In this section, we will explain how we learn different representations, including user representations, item representations, slot \& value representations, and query representations via a unified generative framework. 

\subsubsection{Item Generation Model}
Following the generative process of Equation~\ref{qua:process}, we construct an item generation model. An item $v$ is generated from a user $u$, the user's initial query $Q$, and a sequence of clarifying questions and their associated answers $(q', a')$ which is represented by slot-value pairs $(q, a)$. Inspired by the work of embedding-based generative framework~\cite{mikolov2013distributed,ai2019explainable,bi2019conversational,ai2017learning,zhang2018towards}, we use a softmax function to compute the probability of an item $v$ conditioned by the user embedding, query embedding, and slot-value embeddings:
\begin{equation}
P(v|\mathbf{u},\mathbf{Q},\mathbf{q},\mathbf{a})=
\frac{\text{exp} \big(\mathbf{v} \cdot (\lambda_u \mathbf{u}+\lambda_Q \mathbf{Q}+ \lambda_c \mathbf{c_{qa}}) \big)}{\sum_{\mathbf{v^{\prime}} \in \cal{V}} \text{exp}\big(\mathbf{v^{\prime}} \cdot (\lambda_u \mathbf{u}+\lambda_Q \mathbf{Q}+\lambda_c \mathbf{c_{qa}}) \big)},
\label{equ:item}
\end{equation}
where $\mathbf{c_{qa}}$, $\mathbf{v}$, $\mathbf{u}$, and $\mathbf{Q}$ are the embeddings of $(q, a)$, items, users, and queries, respectively (will be introduced next). $\lambda_u$, $\lambda_Q$, and $\lambda_c$ are hyper-parameters that control the weight of the user, query, and slot-value pairs, respectively. Note that $\lambda_u$ can be 0, which means the model will be degraded to a non-personalized search model. $\lambda_c$ can also be 0, which means the model will be degraded to a non-conversational search model.

\subsubsection{Item and User Language Model}
Inspired by~\citet{ai2017learning}, we learn the representation of items and users by constructing item/user language models.  
While most of the latent vector space models construct item/user language models with textual description (\ie words) only~\cite{ai2017learning,bi2019conversational}, we construct item/user language models from two perspectives, including both textual descriptions and clarification features (\ie slot-value pairs), to learn a better representation of items and users. 

\begin{itemize}
\item When considering textual descriptions, given the item embedding $\mathbf{v}$ ($\mathbf{v} \in \mathbb{R}^t$) and the embedding of a word $\mathbf{w}$ ($\mathbf{w} \in \mathbb{R}^t$), the probability of $w$ being generated from the item language model is defined with a softmax function on item embedding $\mathbf{v}$ and word embedding $\mathbf{w}$:
\begin{equation}
P(w|v)=\frac{\text{exp} (\mathbf{w} \cdot \mathbf{v})}{\sum_{w^{\prime} \in \cal{D}_w} \text{exp} (\mathbf{w^{\prime}} \cdot \mathbf{v} )},
\label{equ:word1}
\end{equation}
where $\cal{D}_w$ is the set of the vocabulary of words in the textual descriptions from the corpus. Similarly, we have a user language model: 
\begin{equation}
P(w|u)=\frac{\text{exp} (\mathbf{w} \cdot \mathbf{u})}{\sum_{w^{\prime} \in \cal{D}_w} \text{exp} (\mathbf{w^{\prime}} \cdot \mathbf{u} )}.
\label{equ:word2}
\end{equation}
\item When considering clarification features, given the item embedding $\mathbf{v}$ ($\mathbf{v} \in \mathbb{R}^t$) and the embedding of a associated slot-value pair $\mathbf{c_{qa}}$ ($\mathbf{c_{qa}} \in \mathbb{R}^t$), the probability of $(q,a)$ being generated from the item language model is defined with a softmax function on the item embedding $\mathbf{v}$ and the slot-value embedding $\mathbf{c_{qa}}$:
\begin{equation}
P\big((q,a)|v\big)=\frac{\text{exp} (\mathbf{c_{qa}} \cdot \mathbf{v})}{\sum_{(q,a)^{\prime} \in \cal{S}_c} \text{exp} (\mathbf{c_{qa}^{\prime}} \cdot \mathbf{v} )},
\label{equ:slot1}
\end{equation}
where $\cal{S}_c$ is the set of clarification features from the corpus. Similarly, we have a user language model: 
\begin{equation}
P\big((q,a)|u\big)=\frac{\text{exp} (\mathbf{c_{qa}} \cdot \mathbf{u})}{\sum_{(q,a)^{\prime} \in \cal{S}_c} \text{exp} (\mathbf{c_{qa}^{\prime}} \cdot \mathbf{u} )}.
\label{equ:slot2}
\end{equation}
\end{itemize}

The item and user language models allow us to make use of the sources of clarification features and textual descriptions, by connecting the embedding of clarification features and the embedding of words with user/item embeddings. This enables us to directly measure the semantic similarity between words and items/users in latent semantic space, as well as the semantic similarity between clarification features and items/users in latent semantic space. 

\subsubsection{Slot-value Representation}
We utilize the informative slot-value pairs $(q, a)$ to represent the conversation, where $q$ represents the slots and $a$ represents the values. In this work, we consider both positive feedback and negative feedback from users. 
We define the representation of slot-value pairs $\mathbf{c_{qa}}$ for positive feedback as:
\begin{equation}
\mathbf{c_{qa}} = \frac{\mathbf{q}+\mathbf{a}}{2},
\end{equation}
where $\mathbf{q}$ and $\mathbf{a}$ are the embedding of the slot in a question and the value in an answer, respectively. For simplicity, we define $\mathbf{c_{qa}}$ as the average of slot and value representations.  Note that the way of combining slot and value representations for $\mathbf{c_{qa}}$ is ad hoc and follows early approaches that pivot on the additive compositionality of embeddings (see \eg \cite{mikolov2013distributed}.); one can define $\mathbf{c_{qa}}$ as any other functions on top of slot and value representations, \eg adding an adaptive weight for the slot and value representations or combining them by a nonlinear projection function~\cite{ai2017learning}.

Besides positive feedback, negative feedback is also a strong signal and is important to be incorporated into the model~\cite{bi2019conversational}. Compared with positive feedback, negative feedback is more challenging since relevant items usually have similar characteristics while the reason for an item to be non-relevant could vary. In this work, we learn a separate embedding for negative feedback. We utilize $\mathbf{q^-}$ to represent the negative feedback on the asked slot $q$.\footnote{Remember that negative feedback on the slot represents the user finding the slot irrelevant to the item she is looking for.} That is, we train both $\mathbf{q}$ and $\mathbf{q^-}$ for each $q$. Then we have the embedding of a clarifying question with negative feedback:
\begin{equation}
\mathbf{c_{qa}} = \mathbf{q^-}.
\end{equation}

\subsubsection{Query Representation}
As there are a large number of possible queries, it is impractical to generalize offline learned query embeddings to unseen queries. That is, $\mathbf{Q}$ must be calculated in product search at the request time. We then divide queries into words and construct query embeddings $\mathbf{Q}$ by word embeddings $\mathbf{w}$ based on a nonlinear projection function similar to other publications~\cite{ai2017learning,van2016learning} :
\begin{equation}
\mathbf{Q}=\text{tanh}(\mathbf{W} \cdot \frac{\sum_{w \in Q} \mathbf{w}}{|Q|} + \mathbf{b}),
\end{equation}
where $\mathbf{W} \in \mathbb{R}^{t \times t}$ and $\mathbf{b} \in \mathbb{R}^{t}$ are parameter matrices and bias vectors to be learned in the training process, respectively. $|Q|$ is the length of the query. Again, the way to combine word embeddings to form a query embedding is ad hoc; one can define query embedding as other functions, \eg aggregating and average the embeddings of query words directly~\cite{vulic2015monolingual}, or sequentially input the query words into a recurrent
neural network (RNN) and use the final network state as the latent query representation~\cite{palangi2016deep}. 

\subsubsection{Joint Learning Framework}
With all the components introduced previously, we can jointly learn the embeddings of queries, users, items, and slot-value pairs by maximizing the likelihood of the observed user-query-conversation-item quadruple in the training set. Let $\cal{S}_{uQv}$ be the set of slot-value pairs in the conversations for the interaction of user $u$ and item $v$ with query $Q$. Let $\cal{S}_u$ be the set of associated slot-value pairs of $u$ (\ie slot-value pairs in the historical conversations for historical purchases of user $u$), $\cal{S}_v$ be the set of associated slot-value pairs of $v$ (\ie slot-value pairs in the historical conversations for historical purchases of item $v$ by all users), $\cal{D}_v$ be the textual descriptions (\ie item description and reviews) of $v$, and $\cal{D}_u$ be the textual descriptions (\ie user reviews) of $u$. Then, the maximization of the likelihood of observing a user-query-conversation-item quadruple with corresponding slot-value pairs and textual descriptions associated with users or items in our model can be formulated as follows:
\begin{equation}
\cal L (u, Q, v, \cal{S}_{uQv}, \cal{S}_v, \cal{S}_u, \cal{D}_v, \cal{D}_u) = \log P(u, Q, v, \cal{S}_{uQv}, \cal{S}_v, \cal{S}_u, \cal{D}_v, \cal{D}_u).
\label{equ:loss}
\end{equation}

In our ConvPS model, words in $\cal{D}_v$ and $\cal{D}_u$ are generated by the language model of
$v$ and $u$, which are Equation~\ref{equ:word1} and Equation~\ref{equ:word2}, respectively. $\cal{D}_v$ is independent of $u$, $Q$, and $\cal{D}_u$ while $\cal{D}_u$ is independent of $v$, $Q$, and $\cal{D}_v$. Slot-value pairs in $\cal{S}_v$ and $\cal{S}_u$ are also generated by the language model of $v$ and $u$, which are Equation~\ref{equ:slot1} and Equation~\ref{equ:slot2}, respectively. $\cal{S}_v$ is independent of $u$, $Q$, $\cal{S}_{uQv}$ and $\cal{S}_u$ while $\cal{S}_u$ is independent of $v$, $Q$, $\cal{S}_{uQv}$ and $\cal{S}_v$. For simplicity, we assume that $\cal{S}_u$, $\cal{S}_v$, and $\cal{S}_{uQv}$ are independent each other. We also assume that $\cal{D}_v$ and $\cal{D}_u$ are independent from $\cal{S}_{uQv}$, $\cal{S}_v$, and $\cal{S}_u$. Thus, we can rewrite Equation~\ref{equ:loss} as follows: 
\begin{equation}
\begin{aligned}
\cal L (u, Q, v, \cal{S}_{uQv}, \cal{S}_v, \cal{S}_u, \cal{D}_v, \cal{D}_u) 
&\propto \sum_{(q,a) \in \cal{S}_{uQv}} \log P(v|u,Q,q,a) + \sum_{(q,a) \in \cal{S}_v} \log  P\big((q,a)|v\big) \\
&+ \sum_{(q,a) \in \cal{S}_u} \log P\big((q,a)|u\big) 
+ \sum_{w \in \cal{D}_v} \log P(w|v)  + \sum_{w \in \cal{D}_u} \log P(w|u).
\end{aligned}
\end{equation}

Therefore, the log-likelihood is actually the sum of log-likelihood for the user language model, the item language model, and the item generation model. To better learn the representations and alleviate the potential mismatch between the query and item for a given user when there are no conversations (\eg the initial item ranking before asking questions), we also add a non-conversational term for user-query-item triples, \ie a non-conversational version of Equation~\ref{equ:item}, into the loss function for optimizing:
\begin{equation}
P(v|u,Q)=\frac{\text{exp} \big(\mathbf{v} \cdot (\lambda_u \mathbf{u}+\lambda_Q \mathbf{Q}) \big)}{\sum_{\mathbf{v^{\prime}} \in \cal{V}} \text{exp}\big(\mathbf{v^{\prime}} \cdot (\lambda_u \mathbf{u}+\lambda_Q \mathbf{Q}) \big)}.
\label{equ:item2}
\end{equation}

Our ablation experiments in Section~\ref{sec:RQ3} also show better performance of involving such a non-conversational term for training. By adding it into our loss function, then we have: 
\begin{equation}
\begin{aligned}
\cal L (u, Q, v, \cal{S}_{uQv}, \cal{S}_v, \cal{S}_u, \cal{D}_v, \cal{D}_u) &= \log P(u, Q, v, \cal{S}_{uQv}, \cal{S}_v, \cal{S}_u, \cal{D}_v, \cal{D}_u)\\
&\propto \log P(v|u,Q) + \sum_{(q,a) \in \cal{S}_{uQv}} \log P(v|u,Q,q,a) + \sum_{(q,a) \in \cal{S}_v} \log  P\big((q,a)|v\big) \\
&+ \sum_{(q,a) \in \cal{S}_u} \log P\big((q,a)|u\big) 
+ \sum_{w \in \cal{D}_v} \log P(w|v)  + \sum_{w \in \cal{D}_u} \log P(w|u).
\end{aligned}
\label{equ:loss2}
\end{equation}

Computing the log-likelihood involving softmax function directly is impractical, since its denominator contains the sum of a large number of elements. Same as other publications~\cite{ai2017learning,levy2014neural}, we approximate the estimation of the denominator of softmax functions by adopting the negative sampling strategy for efficient training. In particular, we randomly sample $\alpha$ words or slot-value pairs from the corpus according to a predefined distribution to approximate the softmax function. Therefore, the log-likelihood of the user language model and item language model with negative sampling is:
\begin{equation}
\begin{aligned}
\log P(v|u,Q) &\approx \log \sigma\big(\mathbf{v} \cdot (\lambda_u \mathbf{u}+\lambda_Q \mathbf{Q}) \big) \\ &+ \alpha \cdot \mathbb{E}_{v^{\prime} \sim P_v}
[\log \sigma\big(-\mathbf{v^{\prime}} \cdot (\lambda_u \mathbf{u}+\lambda_Q \mathbf{Q}) \big)], \\
\log P(v|u,Q,q,a) &\approx \log \sigma\big(\mathbf{v} \cdot (\lambda_u \mathbf{u}+\lambda_Q \mathbf{Q}+ \lambda_c \mathbf{c_{qa}}) \big) \\ 
&+ \alpha \cdot \mathbb{E}_{v^{\prime} \sim P_v}
[\log \sigma\big(-\mathbf{v^{\prime}} \cdot (\lambda_u \mathbf{u}+\lambda_Q \mathbf{Q}+ \lambda_c \mathbf{c_{qa}}) \big)], \\
\log P\big((q,a)|v\big) &\approx \log \sigma(\mathbf{c_{qa}} \cdot \mathbf{v}) + \alpha \cdot \mathbb{E}_{(q,a)^{\prime} \sim P_c}
[\log \sigma(-\mathbf{c_{qa}^{\prime}} \cdot \mathbf{v})], \\
\log P\big((q,a)|u\big) &\approx \log \sigma(\mathbf{c_{qa}} \cdot \mathbf{u}) + \alpha \cdot \mathbb{E}_{(q,a)^{\prime} \sim P_c}
[\log \sigma(-\mathbf{c_{qa}^{\prime}} \cdot \mathbf{u})], \\
\log P(w|v) &\approx \log \sigma(\mathbf{w} \cdot \mathbf{v}) + \alpha \cdot \mathbb{E}_{w' \sim P_w}
[\log \sigma(-\mathbf{w'} \cdot \mathbf{v})], \\
\log P(w|u) &\approx \log \sigma(\mathbf{w} \cdot \mathbf{u}) + \alpha \cdot \mathbb{E}_{w' \sim P_w}
[\log \sigma(-\mathbf{w'} \cdot \mathbf{u})], \\
\end{aligned}
\label{equ:negsample}
\end{equation}
where $P_w$ and $P_c$ is defined as the word distribution and slot-value pair distribution in the corpus respectively, raised to 3/4rd power~\cite{mikolov2013distributed}. $P_v$ is defined as a uniform distribution for items.

By adding L2 regularization to avoid over-fitting, we then define our final loss function:
\begin{equation}
\sum_{u, Q, v} \cal L (u, Q, v, \cal{S}_{uQv}, \cal{S}_v, \cal{S}_u, \cal{D}_v, \cal{D}_u) + \gamma ( \sum_{u \in \cal{U}} \lVert \mathbf{u} \rVert_2^2 +\sum_{v \in \cal{V}} \lVert\mathbf{v}\rVert_2^2 + \sum_{w \in \cal{D}_w} \lVert\mathbf{w}\rVert_2^2 + \sum_{(q,a) \in \cal{S}_c} \lVert\mathbf{c_{qa}}\rVert_2^2),
\end{equation}
where $\gamma$ is the hyper-parameters that control the weight for L2 regularization. All the embeddings are trained simultaneously in our model. 

\subsection{Learning to Ask}
\label{sec:learningtoask}
In this section, we aim to learn how to ask a sequence of high-performance questions. This means, (1) selecting questions to ask to obtain enough information so that the system can suggest the good matching item to the user, and  (2) selecting the questions to ask to hit the target item with as few iterations as possible. In this work, we propose four learning to ask strategies and transfer them to produce search, including \ac{GBS}~\cite{nowak2008generalized}, LinRel bandit~\cite{auer2002using}, and two Gaussian Process (GP) variants. The first one is a greedy strategy while the last three are explore-exploit strategies.

\subsubsection{\ac{GBS}}
Inspired by the successful application in~\citet{zou2020towardsa} and~\citet{zou2020towards}, we explore \ac{GBS} for learning to ask questions in product search. 
\ac{GBS} is a greedy strategy that selects the slot $q_l$ which is able to best split the estimated user preferences corresponding to the items closest to two halves:
\begin{equation}
q_l = \arg\min_{q}\Big|\sum_{v \in \cal{V}} (2 \mathbbm{1}\{q^v = 1\} - 1)\pi_l(v) \Big|,
\label{equ:GBS}
\end{equation}
where $q_l$ is the $l$-th chosen question (slot), $q^v$ expresses whether the textual description of $v$ contains the slot $q$ or not. Specifically, if the slot $q$ appears in the textual description of $v$, then $\{q^v = 1\}$ is true and $\mathbbm{1}\{q^v = 1\}$ is equal to 1, otherwise $\{q^v = 1\}$ is false and $\mathbbm{1}\{q^v = 1\}$ is equal to 0. $\pi_l(v)$ is the estimated user preferences with each dimension in terms of a ranking score of $v$. 
In particular, we set $\pi_l(v)$ to $1/(\text{index}_v +1)$, where $\text{index}_v$ is the index of item $v$ in the $l$-th iteration ranked list of items. 

\subsubsection{LinRel} We also explore bandit algorithms to carry out explore-exploit for learning to ask questions, where we would like to gather information about question rewards through exploration while at the same time harvest currently available question rewards through exploitation. 
Firstly, we adopt a contextual bandit algorithm, LinRel~\cite{auer2002using, ruotsalo2018interactive}, in our task. LinRel has been proven to work well in other interactive retrieval systems~\cite{glowacka2013directing,ruotsalo2018interactive}. LinRel estimates a linear regression model, which makes use of side information (clarification features) to estimate the relevance score of a slot (question). It predicts expected slot relevance and corresponding upper confidence bounds (\ie the sum of the mean and variance of the expected relevance score). To balance the exploration and exploitation, LinRel selects the slot with the highest upper confidence bounds rather than selecting the slot with the largest estimated mean of the relevance score. Intuitively, slots with high upper confidence bounds are the ones that are either already highly relevant with less uncertainty, or potentially relevant but with greater uncertainty. Assume we have a data matrix $\mathbf{X}$ where the slots in terms of questions are rows and items are columns. We denote $\mathbf{X} = (\mathbf{x_1}, \dots, \mathbf{x}_F)^\intercal$ as the matrix containing $F$ row vectors presented so far, where $F$ is the number of slots. 
For any slot $q$, $\mathbf{x_{q}}$ ($\mathbf{x_{q}} \in \mathbf{X}$) is a one-hot vector of items versus that slot $q$ with $N$ dimensions. Each dimension corresponds to a value of 0 or 1 (when a slot is contained in the corresponding item, the value is 1, otherwise 0.). 
The algorithm first computes a regression weight vector for the weight of exploitation (\ie collected question rewards):
\begin{equation}
\mathbf{h_q}=\mathbf{x_{q}} \cdot (\mathbf{X}^\intercal \mathbf{X}+\lambda_I \mathbf{I})^{-1} \mathbf{X}^\intercal,
\label{equ:LinRel1}
\end{equation}
where $\mathbf{I}$ is the identity matrix, and $\lambda_I$ is a regularization parameter.

Suppose up to iteration $l$ we have collected $l$-1 relevance feedback where each feedback
$r_1, \dots, r_{l-1} \in \{-1, 1\}$, is a relevance value for a particular slot, and the vector of all feedback
values received so far is denoted $\mathbf{r} = (r_1, \dots, r_{l-1})^\intercal$. In this paper, we set relevance value of positive feedback to 1 and relevance value of negative feedback to -1. Then we select the slot $q$ that achieve maximum value to ask to the user:
\begin{equation}
q_l=\arg\max_q\big\{\mathbf{h_q} \cdot \mathbf{r} +\frac{c}{2} \lVert\mathbf{h_q}\rVert \big\},
\label{equ:LinRel2}
\end{equation}
where $\mathbf{h_q}$ is the regression weight vector for the slot $q$ whose relevance score we are predicting, $\lVert\mathbf{h_q}\rVert$ is the L2 norm of the regression weight vector, and the constant $c$ is used to adjust the explore-exploit trade-off. Intuitively, the first term on the right side of Equation~\ref{equ:LinRel2} models exploitation and the second term models exploration.

\begin{algorithm}[tb]
\SetKwInOut{Input}{Input}
\Input{Kernel function $\kappa(\cdot,\cdot)$, all of slot representation $\mathbf{x_{q}}$, number of starting points $t$, and number of questions $L$}
\SetKwInOut{Output}{Output}
\Output{The selected slots $(q_{t+1}, \dots, q_L)$: $L>(t+1)$}
Initialization for obtaining $t$ starting points by GBS: 
$\cal{D}_s=\{( q_1, y_1 ), \dots, ( q_t, y_t )\}$\

$l = t+1$\

  \While{$l \leq L$}{
    Update $\mu_l(q)$ and $\sigma^2_l(q)$ according to Equation~\ref{equ:mu} and Equation~\ref{equ:sigma} based on $\cal{D}_s$\ 
    
    select $q_l$ according to Equation~\ref{equ:UCB} or Equation~\ref{equ:EI}\
    
    Observe $y_l \in \{-1, 1\}$ via received feedback of $q_l$\
    
    Adding $(q_l, y_l)$ to $\cal{D}_s$:\ 
    
    \qquad $\cal{D}_s=\cal{D}_s \cup \{(q_l, y_l)\}$\
  
  }
\caption{Gaussian process strategy}
\label{algo1}
\end{algorithm}
\subsubsection{Gaussian Process}
Learning to ask questions can be regarded as making sequential decisions (\ie asking questions) to optimize an unknown payoff function, which can be typically modeled as a Gaussian Process (GP). In particular, each decision (asked question) results in a stochastic reward with an initially unknown distribution, while new decisions are taken on the basis of the observations of previous question rewards. 
In this section, we explore learning to ask strategies modeled by GP. The key idea is to use GP models for the received feedback of asked slots, and to obtain slot estimates via maximum likelihood. In GP, we first deduce the distribution of the relevance of slots and then select the optimal slot to ask based on the relevance score of slots. That is, we predict the relevance score of as yet unexplored slots using GP models, whose regularity is captured in the kernel (covariance) function. Exploiting such regularity allows us to efficiently maximize the benefit drawn from sparse observations about the asked slots. 

Consider now we have already observed relevance feedback for $t$ slots, \ie obtained data $\{( q_1, y_1 ), \dots, ( q_t, y_t )\}$, where $y_t$ is the observed relevance of $q_t$. In this work, $y_t \in \{-1, 1\}$, \ie it is set to 1 for positive feedback received and -1 for negative feedback received. We assume the relevance of slot $y_t$ conforms to a GP. We also assume $y_t$ is noisy and is composed of a ``true'' relevance $R(q_t)$ plus independent Gaussian noise $e_t$, with $y_t = R(q_t) + e_t$, where $e_t$ is a predefined hyper-parameter. The GP assumption for the observed feedback $(y_1,\dots,y_t)$ implies that the true feedback for all slots are jointly Gaussian distributed, with zero mean. Then, for a new slot, its predictive distribution for the relevance $R(q)$ is Gaussian, with mean and variance given by
\begin{equation}
\mu_t(q)= \mathbf{k}_t(\mathbf{x_q})^\intercal (\mathbf{K}_t+\mathbf{I})^{-1} \mathbf{y}_t,
\label{equ:mu}
\end{equation}
\begin{equation}
\sigma^2_t(q) = \kappa(\mathbf{x_q},\mathbf{x_q})- \mathbf{k}_t(\mathbf{x_q})^\intercal (\mathbf{K}_t+\mathbf{I})^{-1} \mathbf{k}_t(\mathbf{x_q}),
\label{equ:sigma}
\end{equation}
where $\mathbf{k}_t(\mathbf{x_q}) = [\kappa (\mathbf{x_q},\mathbf{x_{q_1}} ), \dots, \kappa( \mathbf{x_q},\mathbf{x_{q_t}} )]^T$, $\mathbf{K}_t$ is the positive semi-definite kernel matrix such that $\mathbf{K}_{t,i,j} = [\kappa(\mathbf{x_{q_i}},\mathbf{x_{q_j}})]$, and $\kappa(\mathbf{x_{q_i}}, \mathbf{x_{q_j}})$ is a particular kernel function for two embeddings $\mathbf{x_{q_i}}$ and $\mathbf{x_{q_j}}$. Again, $\mathbf{x_{q_i}}$ and $\mathbf{x_{q_j}}$ are one-hot vectors of items versus the slot $q_i$ and $q_j$ respectively, with each dimension corresponding to a value of 0 or 1. For two slots $q_i$ and $q_j$, the kernel $\kappa(\mathbf{x_{q_i}}, \mathbf{x_{q_j}})$ represents our assumptions about how similar we expect the relevance of $q_i$ with representation $\mathbf{x_{q_i}}$, as opposed to $q_j$ with representation $\mathbf{x_{q_j}}$ are. In this work, we use a common kernel function: RBF kernel $\kappa(\mathbf{x_{q_i}},\mathbf{x_{q_j}})=\sigma^2 \text{exp} (-\frac{||\mathbf{x_{q_i}} - \mathbf{x_{q_j}}||^2}{2})$. RBF kernel is a popular kernel function used in Support Vector Machine (SVM)~\cite{suthaharan2016support} and various kernelized learning algorithms~\cite{li2021crowdgp}. 

\paragraph{Choice of acquisition function} For GP, we need an acquisition function for deciding where to sample slots. In this work, we use two acquisition functions to select slots. 
\begin{itemize}
    \item Upper Confidence Bound (UCB): 
\begin{equation}
q_l=\arg\max_{q} \mu_{l}(q) + \beta \sigma_{l}(q).
\label{equ:UCB}
\end{equation}
This criterion selects the slot with the highest upper confidence bound on the relevance function. $\beta$ is a tradeoff parameter to control the explore-exploit.
    \item Expected Improvement (EI):
\begin{equation}
q_l = \arg\max_{q} \text{EI}_{l}(q), 
\label{equ:EI}
\end{equation}
\begin{equation}
\text{EI}_{l}(q) =(\mu_{l}(q) - \mu_{l}(q_l^*)) \Phi \frac{\mu_{l}(q) - \mu_{l}(q_l^*)}{\sigma_{l}(q)} + \sigma_{l}(q) \phi \frac{\mu_{l}(q) - \mu_{l}(q_l^*)}{\sigma_{l}(q)},
\label{equ:EI2}
\end{equation}
where $\phi$ and $\Phi$ are the standard normal probability density and cumulative distribution functions, respectively. $q_l^* = \arg\max_{q}\mu_{l}(q)$, which denotes the slot with largest posterior mean at round $l$. The EI value $\text{EI}_{l}(q)$ measures the potential of $q$ to improve upon the largest posterior mean $\mu_{l}(q_l^*)$ at round $l$. EI is another popular infill criterion in GP assisted optimization to balance exploitation and exploration of the predictive distribution model, which has been widely used~\cite{jiao2019complete}. It selects the slot with the largest amount of improvement $\text{EI}_{l}(q)$.
\end{itemize}

The two acquisition functions yield two GP strategies, including EI-based GP strategy (GP+EI) and UCB-based GP strategy (GP+UCB), which are shown in Algorithm~\ref{algo1}. We first initialize GP by some starting points. 
In this work, we initialize the first two starting points by using GBS, \ie $\{( q_1, y_1 ), ( q_2, y_2 )\}$ is obtained by using GBS. One can also use other techniques (\eg random) to obtain starting points. After the initialization, we update $\mu_l(q)$ and $\sigma^2_l(q)$ according to Equation~\ref{equ:mu} and Equation~\ref{equ:sigma} based on the set of observed relevance feedback. Then we utilize the acquisition function to select a slot $q_l$ according to Equation~\ref{equ:UCB} or Equation~\ref{equ:EI}. After that, we ask the question about $q_l$ and observe the user relevance feedback $y_l$, and then add $(q_l, y_l)$ to the set of observed relevance feedback to update $\mu_l(q)$ and $\sigma^2_l(q)$ again.

\subsection{Item Ranking with ConvPS}
After we train the embeddings of users, items, words, and slot-value pairs, we can retrieve items based on that. When a user $u$ issues an initial
query $Q$, our system first ranks an item $v$ based on $P(v|u,Q)$ according to Equation~\ref{equ:item2}, in the initial iteration. After $l$-th round clarifying questions are selected to ask to the user, which is demonstrated as $\cal{S}_{qa} = \{(q_1, a_1), (q_2, a_2), \dots, (q_l, a_l)\}$, we rank an item $v$ in the $l$-th iteration ($l>0$) according to: 
\begin{equation}
\label{equ:24}
\begin{aligned}
P(v|u,Q,\cal{S}_{qa})=\frac{\text{exp} \big(\mathbf{v} \cdot (\lambda_u \mathbf{u}+\lambda_Q \mathbf{Q}+ \lambda_c \sum_{(q,a) \in \cal{S}_{qa}} \mathbf{c_{qa}}) \big)}{\sum_{\mathbf{v^{\prime}} \in \cal{V}} \text{exp}\big(\mathbf{v^{\prime}} \cdot (\lambda_u \mathbf{u}+\lambda_Q \mathbf{Q}+\lambda_c \sum_{(q,a) \in \cal{S}_{qa}} \mathbf{c_{qa}}) \big)}.
\end{aligned}
\end{equation}

\section{Experiments and Analysis}
\label{sec:exp}
In this section, we start by introducing our experimental setup, including the datasets, evaluation metrics, baselines, conversation construction, parameter settings, and our research questions. We then demonstrate and analyze the results in terms of each of our research questions.

\subsection{Experimental Setup}
\label{ExpSetup}
\subsubsection{Datasets.} 
For our experiments, we use the Amazon product datasets\footnote{http://jmcauley.ucsd.edu/data/amazon/}~\cite{mcauley2015image}, as in previous product search publications~\cite{van2016learning, ai2017learning,zhang2018towards,bi2019conversational}. 
The datasets contain millions of users and items. Each item contains rich metadata such as item descriptions, item categories, and user reviews. In this paper, we use three item categories with diverse scales from the Amazon product datasets, which are ``Movies \& TV'', ``Health \& Personal Care'', and ``Cell Phones \& Accessories'', following ~\citet{bi2019conversational}. The former one constitutes a very large dataset while the rest are smaller, which allows us to test the effectiveness of our ConvPS model on collections of different sizes. 
The training/test set splitting for the three corresponding item categories is also the same with~\citet{bi2019conversational}. As there is no publicly available product search dataset containing search queries, we construct initial queries following the same paradigm employed in~\citet{ai2017learning},~\citet{bi2019conversational},~\citet{van2016learning},~\citet{zhang2018towards}, and~\citet{zhang2018towards}, \ie we use a subcategory title from the above three item categories to form a topic string as a query. Each subcategory (\ie query) includes multiple (relevant) items and each item can be relevant to multiple queries (subcategories). We evaluate the model performance for each user-query pair, \ie each individual user purchases belonging to the initial sub-category (query) observed in the data. If an item purchased by a user belongs to multiple sub-categories (queries), the pairing of this user with any of these queries creates valid user-query pairs. For each user-query pair, we rank all possible items within each individual category dataset. Therefore, the number of candidate items is very large and target items for each user-query pair are very sparse, which leads to a difficult product search setting by nature. 
The statistics for our datasets are shown in Table~\ref{tab:dataset}. 
\begin{table}[t]
\caption{Basic Statistics of our experimental datasets. Tr and Te denote the train and test sets, respectively. Arithmetic mean and standard deviation are indicated wherever applicable.}
\label{tab:dataset}
\centering
%  \small
\begin{tabular}{p{0.25\columnwidth}|p{0.2\columnwidth}<{\centering}|p{0.2\columnwidth}<{\centering}|p{0.2\columnwidth}<{\centering}}
% \begin{tabular}{l|c|c|c}
\toprule
 & \textbf{Cell Phones \& Accessories} & \textbf{Health \& Personal Care}  & \textbf{Movies \& TV}\\
 \hline
 \# Users & 27,879 & 38,609 & 123,960\\
 \# Items & 10,429 & 18,534 & 50,052 \\
 \# Reviews & 194,439 & 346,355 & 1,697,524 \\
 \# Queries & 165 & 779  & 248\\
 \# Words per query& 5.93 $\pm$ 1.57 & 8.25 $\pm$ 2.16  & 5.31 $\pm$ 1.61 \\
 \hdashline
 \# Slots (questions) & 738 & 1,906  & 6,694\\
 \# Slot-value pairs & 7,849 & 17,203 & 88,754\\
\hdashline
%Train/Test &&&&\\
 \# User-query pairs & 114,177 (Tr) & 231,186 (Tr) & 241,436 (Tr)\\
 & 665 (Te) & 282 (Te) & 5,209 (Te)\\
 \# Relevant items per pair & 1.52 $\pm$ 1.13 (Tr) & 1.14 $\pm$ 0.48 (Tr) & 5.40 $\pm$ 18.39 (Tr) \\
 & 1.00 $\pm$ 0.05 (Te) & 1.00 $\pm$ 0.00 (Te) & 1.10 $\pm$ 0.49 (Te) \\ 
\bottomrule
\end{tabular}
\end{table}

\subsubsection{Evaluation Metrics.}
As in~\citet{ai2017learning},~\citet{bi2019conversational}, and~\citet{zhang2018towards}, we use mean average precision (MAP), mean reciprocal rank (MRR) and normalized discounted cumulative gain (NDCG) as our evaluation metrics, where MAP and MRR are calculated based on top 100 items retrieved by each model, and NDCG is calculated based on top 10 items. Intuitively, MRR indicates the expected number of items that a user needs to browse before finding the first relevant item to purchase. 
MAP measures the overall performance taking both precision and recall into account. 
NDCG measures the ranking performance of results at the top of the ranking compared to an optimal ranked list. We are expecting the three measures to be strongly correlated. 

\subsubsection{Baselines.} The following baselines are used:
\label{baseline}
\begin{itemize}
    \item \textbf{LSE}%\footnote{https://github.com/cvangysel/SERT}
    ~\cite{van2016learning}, which is the first latent space model proposed for product search. 
    \item \textbf{HEM}%\footnote{https://github.com/QingyaoAi/Hierarchical-Embedding-Model-for-Personalized-Product-Search}
    ~\cite{ai2017learning}, which is a state-of-the-art retrieval model for personalized product search;
    \item \textbf{PMMN}%\footnote{}
    ~\cite{zhang2018towards}, which is a state-of-the-art conversational search system asking questions on slot-value pairs. 
    \item \textbf{AVLEM}%\footnote{https://github.com/kepingbi/ConvProductSearchNF}
    ~\cite{bi2019conversational}, which is a state-of-the-art conversational product search model by incorporating negative feedback. 
\end{itemize}
The first two baselines are two typical static baselines, which are widely used~\cite{bi2019conversational,zhang2018towards}. The last two baselines are two state-of-the-art conversational baselines. We only include neural models on the basis of learning representations for soft-matching as our baselines since exact term matching has been proven insufficient to retrieve ideal products and term-based retrieval models are less effective in previous studies on product search~\cite{bi2019conversational, ai2017learning}.

\subsubsection{Conversation Construction.} 
With the absence of the conversational product search dataset, we simulate users and construct conversations based on the Amazon product dataset, as done in~\citet{bi2019conversational}, ~\citet{zhang2018towards}, and~\citet{zou2019learning}. We also
conduct a small user study by involving real users described in Section~\ref{user_study}. 
% \begin{enumerate}[(a)]
Specifically, we simulate conversations by adopting the approach of ~\citet{bi2019conversational} and~\citet{zhang2018towards}, which extracts slot-value pairs from reviews of items by using aspect-value pair extraction toolkit~\cite{zhang2014explicit}. 
During training, all relevant (positive) slots for a user-item pair $(u,v)$ are used to construct the questions. 
In order to train representations of slots with negative feedback, we also randomly select non-relevant (negative) slots from the entire slot pool for a user-item pair $(u,v)$ to construct the questions. 
During testing time, we select the slots to form corresponding questions from the entire slot pool constructed by the training set, by using our proposed question selection strategies. We will never ask the slots that do not appear in the training set. 
For the answer during training and testing time, we simulate user feedback following previous literature~\cite{bi2019conversational,zhang2018towards,sun2018conversational}. They construct users' responses to the asked questions according to users' target items, which show their hidden intents. The system asks users their preferred value of a slot and answers are constructed according to the corresponding value for that slot in their target items. 
Therefore, same with them~\cite{bi2019conversational,zhang2018towards,sun2018conversational}, we assume that a user has target items in mind. 
For an asked question on the basis of a slot, the users will respond with the corresponding value for that slot in their target items when the slot is contained in the target items, while responding with negative feedback ``not relevant'' when a slot is absent from the target items. 
In the case that the simulated user answers in the testing set are not in the training set, therefore leading to an embedding mismatching, we regard it as an invalid question and ignore it for updating the ranking results. 
Unlike~\cite{bi2019conversational,zhang2018towards} only ask slots appearing in the target items, we ask slots from the entire slot pool (slots appearing in the entire item set), what we believe are not exclusive and more reasonable, as well as more challenging by nature. 

\subsubsection{Online User Study.} 
\label{user_study}
Besides the simulated users, we develop an online Web system to conduct a small user study by involving real users. The purpose of the online user study is to validate some of assumptions in our conversational product search system and test the performance of our system ``in-situ''. 
Ideally, the existing customers of an online shopping platform who have previously purchased a number of products and now interact with our conversational product search system embedded in the shopping platform to locate their next target products would be optimal. In the absence of such a commercial product search system and user base, we utilize a crowdsourcing platform\footnote{https://www.prolific.co/} to recruit crowd workers to interact with our online Web system. We first ask the crowd worker to choose a product category he/she feels comfortable with from our dataset. Then, we randomly sample a product from our test data in the selected product category as a target product. The user is then guided to carefully review the information about the target product, \eg a product image, title, description, and the extracted slot-value pairs.  
Once the user indicates that he/she is familiar with the target product, the target product disappears from the screen and the conversation with the system starts. Then, a question selected by our algorithm is asked to the user. In this user study, we use our ConvPS algorithm with LinRel question selection strategy. After the user observes the algorithmically chosen question, he/she can provide an answer to the question according to the information of the target product. The user has the option to stop answering questions at any time during her interaction with the system. When the interaction with the system stops, we ask users a series of exit questions regarding their experiences.

\subsubsection{Parameter Settings.} 
We set the hyper-parameters to the combination that achieved the highest pairwise accuracy in the offline observations as follows: we train our ConvPS model with stochastic gradient descent for 20 epochs with batch size 64. For stochastic gradient descent in training, the initial learning rate is set as 0.5 and gradually decreases to 0. We use the global norm clip of gradients with 5 for stable training. We set hyper-parameters $\lambda_I$ = $e_i$ = 0.1, and $\lambda_u$ = $\lambda_Q$ = $\lambda_c$ = 1. 
The number of negative samples in Equation~\ref{equ:negsample} is set as 5 and L2 regularization strength $\gamma$ is tuned from 0 to 0.01. 
As in~\citet{ai2017learning}, and~\citet{bi2019conversational}, we set the subsampling rate of words for ``Movies \& TV'', and ``Cell Phones \& Accessories'' as $10^{-6}$ and $10^{-5}$, respectively, to speed up the training and reduce the effect of common words. We study the impact of the number of questions asked, the embedding size, the batch size, and the explore-exploit trade-off parameter $c$ in Section~\ref{sec:parameter}. The number of questions asked for the conversation is examined from 0 to 10 with step size 1, the embedding size is examined from 100 to 500 with step size 100, the batch size is tuned from [64, 128, 256, 612], the explore-exploit trade-off parameters $c$ and $\beta$ are tested from [0, 2, 4, 6, 8, 10]. Unless mentioned otherwise, we use 5 questions and other optimal parameters, which are 200 for the embedding size, 64 for the batch size, and 4, 2 for the explore-exploit trade-off parameter $c$ and $\beta$ respectively, to report results. For the parameter settings of the four baselines, we use the optimal parameters stated in the corresponding product search publications and tuned their hyper-parameters in the same way as they reported. 

\subsubsection{Research Questions.} 
In our experiments, we will investigate and answer the following four questions:
\RQ{1}{How effective is our proposed model ConvPS compared to prior works?}
\RQ{2}{Which question selection strategy is the best?}
\RQ{3}{What are the effects of different model components?}
\RQ{4}{How do the parameters of ConvPS affect its efficacy?}

\begin{table}[t]
\caption{Performance comparison between our ConvPS model and baselines (LSE, HEM, PMMN, and AVLEM) on the ``Cell Phones \& Accessories'', ``Health \& Personal Care'', and ``Movies \& TV'' categories. The performance of PMMN, AVLEM and our ConvPS model are reported when 5 questions (slots) are asked. `*' indicates significant differences between ConvNF and $\text{ConvPS}_{init}$ in Fisher random test with $p\text{-value} < 0.05$,  `$\dagger$' denotes significant improvements upon the best baseline with $p\text{-value} < 0.05$. Best performances are in bold.}
\label{table:compare}
\centering
\small
\setlength{\tabcolsep}{0.8\tabcolsep}
\begin{tabular}{l|ccc|ccc|ccc}
\toprule
& \multicolumn{3}{c|}{Cell Phones \& Accessories} & \multicolumn{3}{c|}{Health \& Personal Care} & \multicolumn{3}{c}{Movies \& TV}\\
\hline
Model & MAP & MRR & NDCG & MAP & MRR & NDCG & MAP & MRR & NDCG\\
 \hline
LSE & 0.098 & 0.098 & 0.084 & 0.155 & 0.157 & 0.195 & 0.023 & 0.025 & 0.027 \\
HEM & 0.115 & 0.115 & 0.116 & 0.189 & 0.189 & 0.201 & 0.026 & 0.030 & 0.030 \\
\hdashline
PMMN & 0.115 & 0.116 & 0.122 & 0.180 &0.180 &0.196 & 0.029 & 0.030 & 0.023 \\
AVLEM & 0.125 & 0.125 & 0.130 & 0.214 & 0.214 & 0.251 & 0.025 & 0.026 & 0.027 \\
\hdashline
$\text{ConvPS}_{init}$ & 0.126 & 0.126 & 0.136 &0.243 &0.243 & 0.267 & 0.021 & 0.023 & 0.024\\
\hdashline
$\text{ConvPS}_{random}$ & 0.131*$\dagger$ & 0.131*$\dagger$ & 0.141*$\dagger$ & 0.254*$\dagger$ &0.254*$\dagger$ &0.279*$\dagger$ & 0.022 & 0.023 & 0.024 \\
$\text{ConvPS}_{GBS}$  & 0.226*$\dagger$ & 0.226*$\dagger$ & 0.258*$\dagger$ & 0.340*$\dagger$ & 0.340*$\dagger$ & 0.359*$\dagger$ & 0.042*$\dagger$ & 0.045*$\dagger$ & 0.048*$\dagger$\\
$\text{ConvPS}_{LinRel}$  & \textbf{0.235}*$\dagger$ & \textbf{0.236}*$\dagger$ & \textbf{0.265}*$\dagger$ & 0.401*$\dagger$ &0.401*$\dagger$ & 0.413*$\dagger$ & \textbf{0.044}*$\dagger$ & \textbf{0.046}*$\dagger$ & \textbf{0.050}*$\dagger$\\
$\text{ConvPS}_{GP+EI}$  & 0.223*$\dagger$ & 0.223*$\dagger$ & 0.242*$\dagger$ &\textbf{0.408}*$\dagger$  & \textbf{0.408}*$\dagger$ & \textbf{0.441}*$\dagger$ & 0.035*$\dagger$ & 0.037*$\dagger$ & 0.039*$\dagger$\\
$\text{ConvPS}_{GP+UCB}$  & 0.231*$\dagger$ & 0.231*$\dagger$ & 0.251*$\dagger$ & 0.347*$\dagger$ & 0.347*$\dagger$ & 0.376*$\dagger$ & 0.035*$\dagger$ & 0.038*$\dagger$ & 0.040*$\dagger$ \\
\bottomrule
\end{tabular}
\end{table}
\subsection{RQ1 Overall Performance Comparison}
To study the overall effectiveness of our ConvPS model, we compare the performance of our ConvPS, measured by MAP, MRR, and NDCG, with baselines shown in Section~\ref{baseline}. The results on the ``Cell Phones \& Accessories'', ``Health \& Personal Care'', and ``Movies \& TV'' categories are shown in Table~\ref{table:compare}. We refer to the ConvPS with random question selection, with GBS question selection strategy, with LinRel question selection strategy, with EI-based GP question selection strategy, and with UCB-based GP question selection strategy as  $\text{ConvPS}_{random}$, $\text{ConvPS}_{GBS}$, $\text{ConvPS}_{LinRel}$, $\text{ConvPS}_{GP+EI}$, and $\text{ConvPS}_{GP+UCB}$, respectively. 

As shown in Table~\ref{table:compare}, we observe that our ConvPS model with different question selection strategies ($\text{ConvPS}_{random}$, $\text{ConvPS}_{GBS}$, $\text{ConvPS}_{LinRel}$, $\text{ConvPS}_{GP+EI}$, and $\text{ConvPS}_{GP+UCB}$) outperforms the baselines LSE, HEM, PMMN, and AVLEM, on all three metrics, which suggests that our ConvPS model is effective. As expected, with 5 questions asked, our ConvPS model with different question selection strategies significantly outperforms the static baselines LSE and HEM. We attribute this to the reason that our ConvPS can continuously learn from the user and clear users' item needs through conversations. With 0 questions asked, our $\text{ConvPS}_{init}$ model can still achieve comparable performance on ``Movies \& TV'' and better performance on ``Cell Phones \& Accessories'' and ``Health \& Personal Care''. 
Further, our ConvPS models perform better than the two dynamic baselines PMMN and AVLEM. This suggests that our ConvPS models better learn user true preferences. This might be explained by the fact that the embeddings (\ie user embeddings, query embeddings, item embeddings, and conversation embeddings in terms of slot-value embeddings) learned by our ConvPS models can better distinguish the target items from those irrelevant items, or our question selection strategies can better gather informative user feedback. 
Compared with $\text{ConvPS}_{init}$, we observe consistent and large improvements of our ConvPS model with the four different question selection strategies (\ie $\text{ConvPS}_{GBS}$, $\text{ConvPS}_{LinRel}$, $\text{ConvPS}_{GP+EI}$, and $\text{ConvPS}_{GP+UCB}$), which is approximately 100\% in terms of three metrics in all three categories. This clearly indicates the contribution of conversations, suggesting conversational product search is highly beneficial. 

\subsection{RQ2 Question Selection Strategy Comparison}

\begin{figure}[t]
\begin{subfigure}[t]{0.32\textwidth}
\includegraphics[width=\columnwidth]{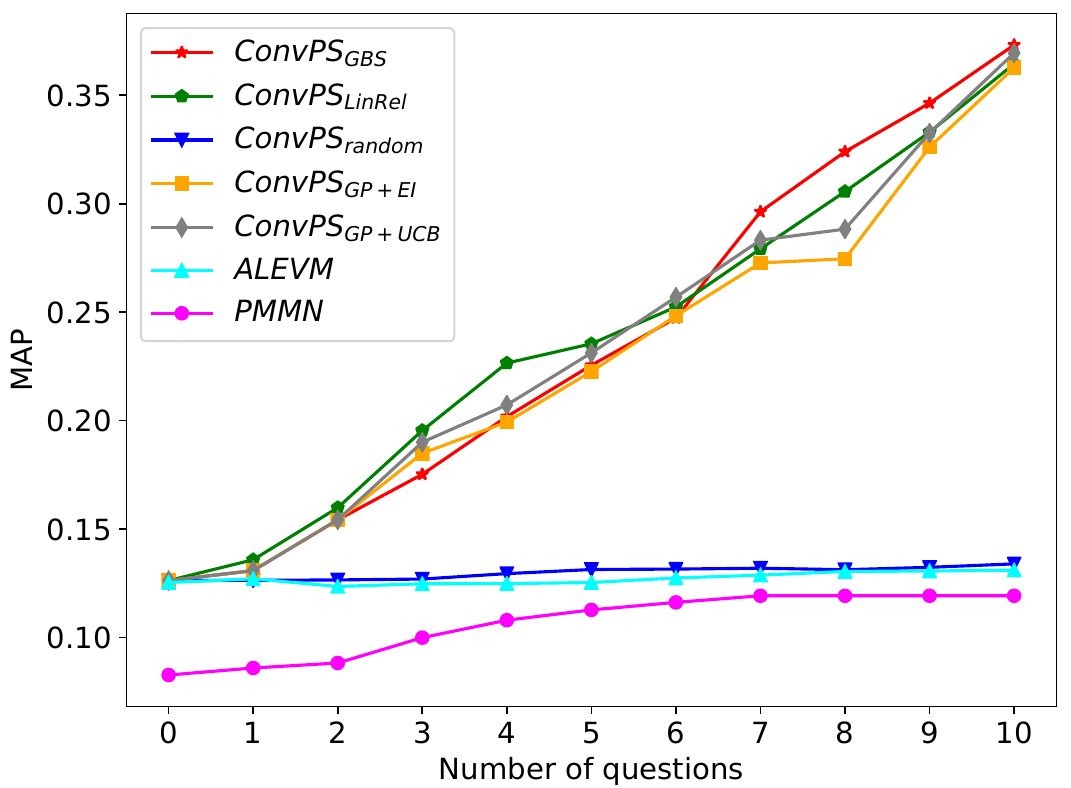}
  \caption{MAP}
  \label{fig:Questions_MAP}
\end{subfigure}
\begin{subfigure}[t]{0.32\textwidth}
  \includegraphics[width=\columnwidth]{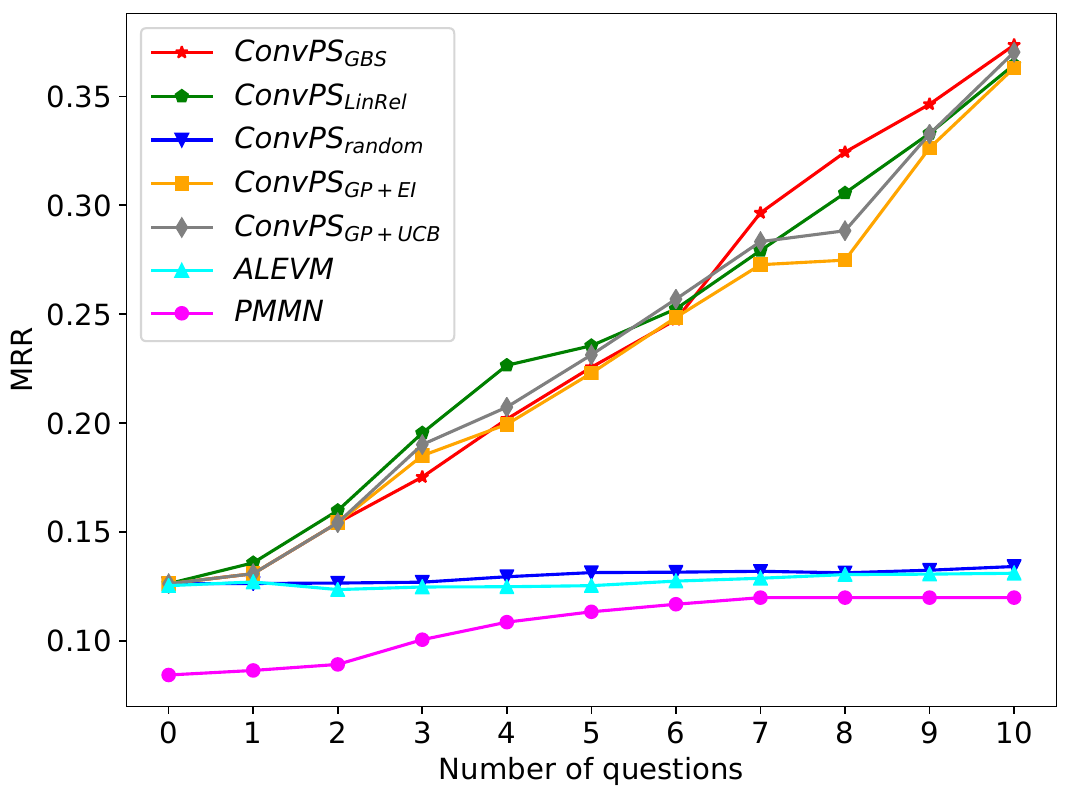}
  \caption{MRR}
  \label{fig:Questions_MRR}
\end{subfigure}
\begin{subfigure}[t]{0.32\textwidth}
  \includegraphics[width=\columnwidth]{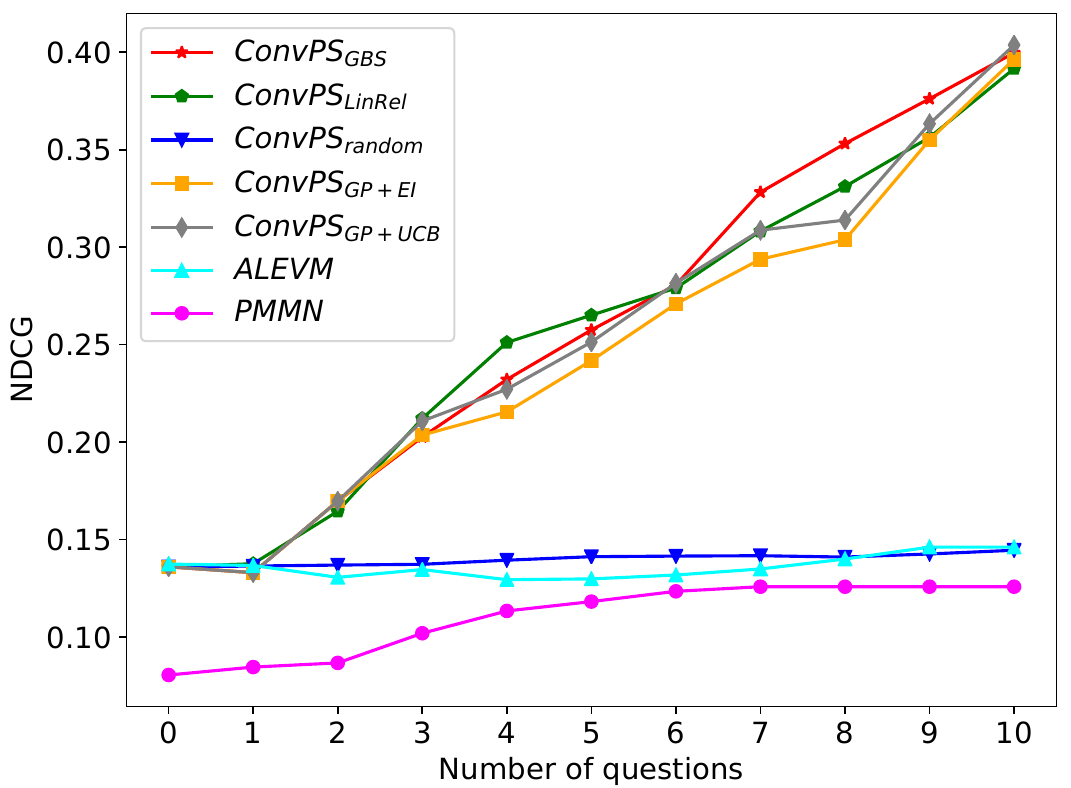}
  \caption{NDCG}
  \label{fig:Questions_NDCG}
\end{subfigure}
    \caption{Performance of different question selection strategies.}
     \label{fig:question}
\end{figure}

In Figure~\ref{fig:question}, we report the comparison results of the various question selection strategies on ``Cell Phones \& Accessories''. We see that 
our ConvPS models outperform PMMN and AVLEM and the performance gains become larger as the number of questions increases. What we indeed
observed is that the performance on all three metrics regarding PMMN and AVLEM increases slowly while the performance of our ConvPS models improves greatly, as the number of questions increases. 
This again demonstrates that our ConvPS models learn more effectively and efficiently from the user than PMMN and AVLEM, reflected in either representation learning or question selection strategies. 

$\text{ConvPS}_{GBS}$, $\text{ConvPS}_{LinRel}$, $\text{ConvPS}_{GP+EI}$, and $\text{ConvPS}_{GP+UCB}$ achieve much higher performance than $\text{ConvPS}_{random}$, demonstrating our four question selection strategies are highly beneficial. $\text{ConvPS}_{random}$ learns slowly, because it fails to select more promising questions to ask for effective learning from the user. Overall, $\text{ConvPS}_{GBS}$, $\text{ConvPS}_{LinRel}$, $\text{ConvPS}_{GP+EI}$, and $\text{ConvPS}_{GP+UCB}$ perform well and achieve comparable performance. Specifically, we find that the bandit-inspired strategies $\text{ConvPS}_{LinRel}$ and $\text{ConvPS}_{GP+UCB}$ perform slightly better at early stage (\eg less than 6 questions). The plausible reason is that $\text{ConvPS}_{LinRel}$ and $\text{ConvPS}_{GP+UCB}$ systematically balance the need to explore new solutions with the need to reap the rewards of what has already been learned. At latter stage (\eg more than 6 questions), $\text{ConvPS}_{GBS}$ performs slightly better. This might be because the GBS question selection strategy can learn from user preference over items $\pi_l(v)$ as the number of questions increases. With the user preference more close to the true user preference at the latter stage, GBS learns faster than other question selection strategies. 

\begin{table}[t]
\caption{Question selection strategy comparison in terms of the ratio of questions with positive feedback, negative feedback, or invalid feedback.}
\label{table:ratiocompare}
\centering
\begin{tabular}{l|ccc}
\toprule
Model & \% Positive & \% Negative & \% Invalid \\
 \hline
$\text{ConvPS}_{random}$ & 5.0\% & 0.9\% & 94.1\% \\
$\text{ConvPS}_{GBS}$  & 69.6\% & 15.1\% & 15.3\% \\
$\text{ConvPS}_{LinRel}$ & 71.0\% & 26.0\% & 3.0\%\\
$\text{ConvPS}_{GP+EI}$  & 64.1\% & 11.0\% & 24.9\% \\
$\text{ConvPS}_{GP+UCB}$  & 64.0\% & 11.1\% & 24.9\% \\
\bottomrule
\end{tabular}
\end{table}
We further explore why those question selection strategies perform differently by reporting the ratio of questions with positive feedback, negative feedback, or invalid feedback. Ideally, the user is more engaged to be asked those questions with positive feedback, while less engaged be asked questions with negative or invalid feedback. Questions with negative feedback usually mean those questions are irrelevant. 
Due to the great diversity in questions and items, most questions are irrelevant to a target item and they should not be asked to dissatisfy users. Questions with invalid feedback mean those questions are useless leading to no gain. 
As shown in Table~\ref{table:ratiocompare}, we observe that $\text{ConvPS}_{LinRel}$ achieve top performance across models, with the highest ratio of questions with positive feedback, and the lowest ratio of questions with invalid feedback. $\text{ConvPS}_{random}$ asks a large number of questions with invalid feedback, leading to the lowest performance in Figure~\ref{fig:question}. This is because the question pool is large and the percentage of questions with positive feedback in the question pool for a certain target item is low. 

\begin{figure}[t]
\begin{subfigure}[t]{0.32\textwidth}
\includegraphics[width=\columnwidth]{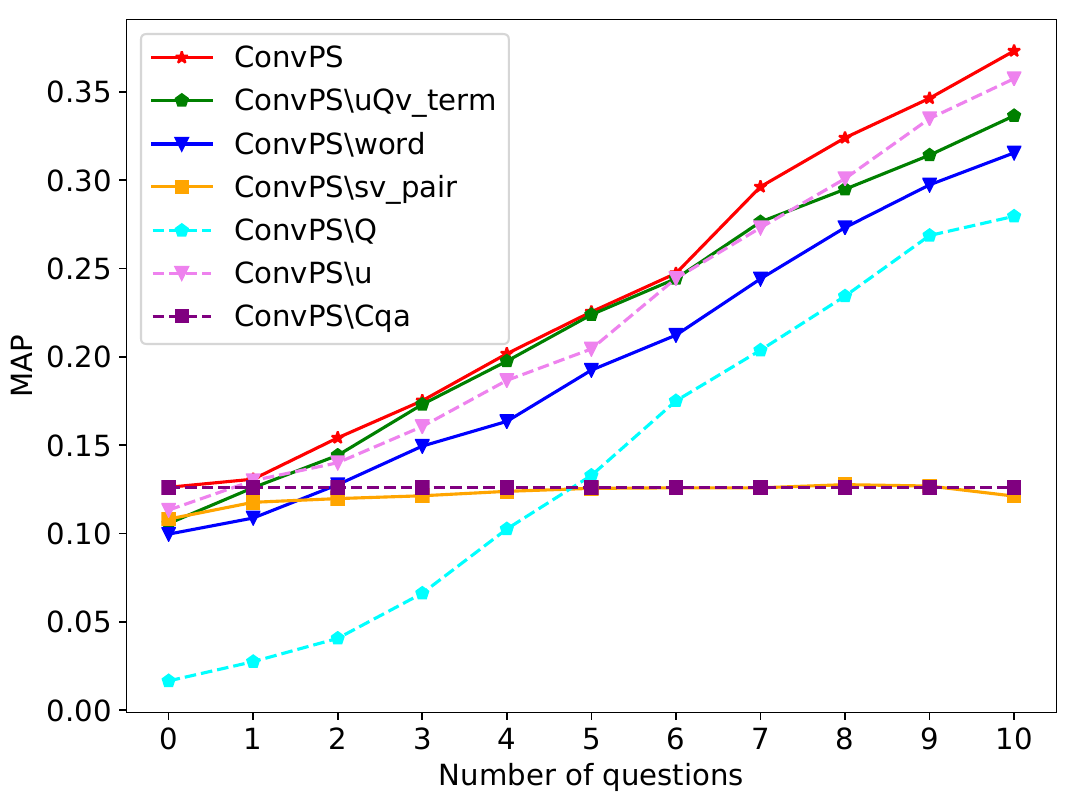}
  \caption{MAP}
  \label{fig:ablation_MAP}
\end{subfigure}
\begin{subfigure}[t]{0.32\textwidth}
  \includegraphics[width=\columnwidth]{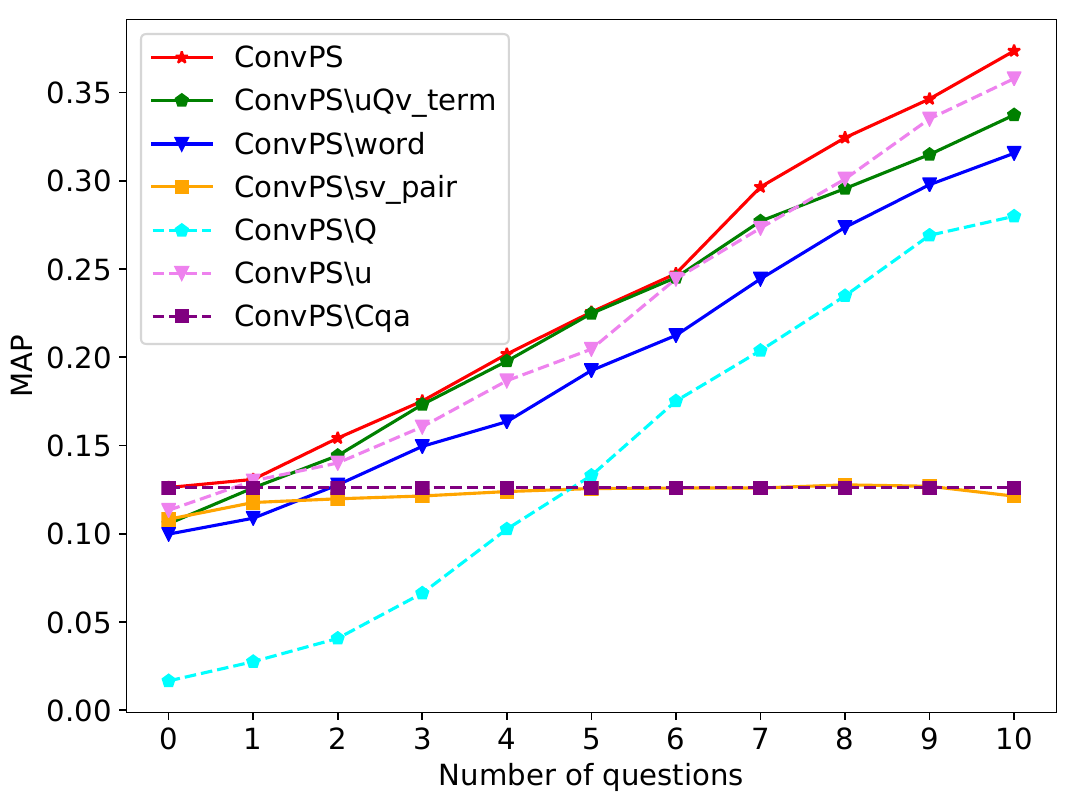}
  \caption{MRR}
  \label{fig:ablation_MRR}
\end{subfigure}
\begin{subfigure}[t]{0.32\textwidth}
  \includegraphics[width=\columnwidth]{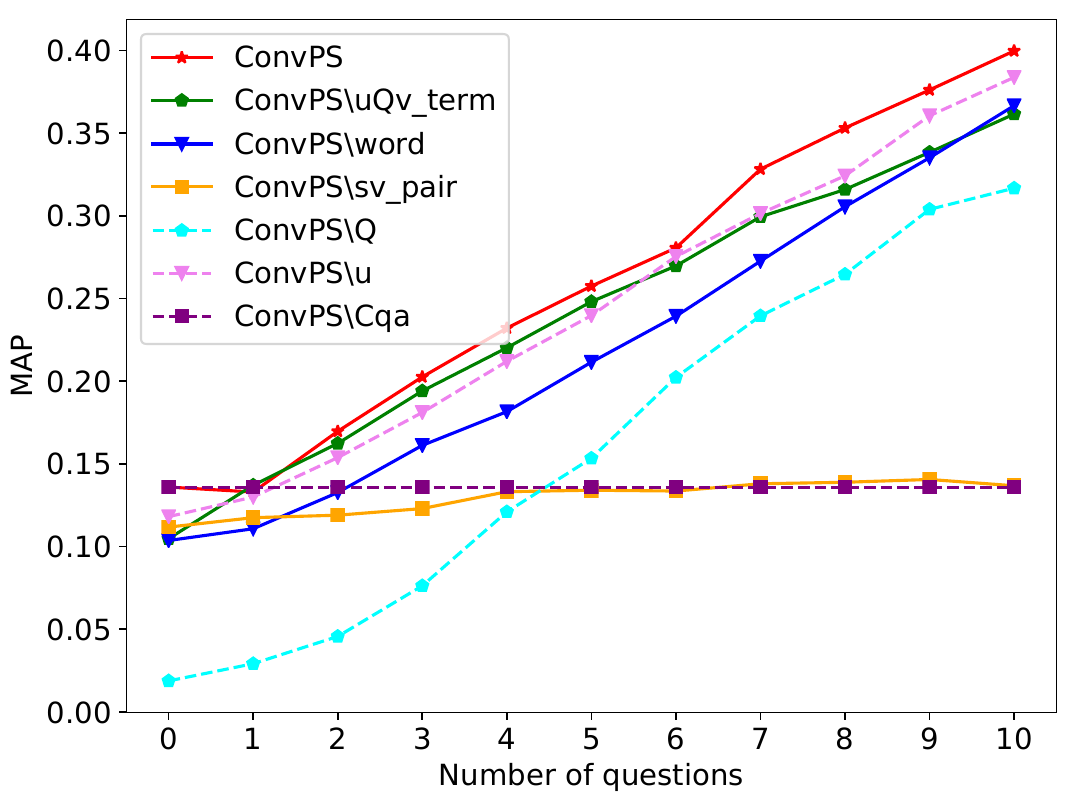}
  \caption{NDCG}
  \label{fig:ablation_NDCG}
\end{subfigure}
    \caption{The performance of ConvPS with different components removed.}
     \label{fig:ablation}
\end{figure}
\subsection{RQ3 Ablation Study}
\label{sec:RQ3}
To evaluate the importance of different components of our ConvPS, we conduct a series of ablation experiments. We refer the ConvPS by removing the user and item language models for textual descriptions (\ie removing the corresponding terms of Equation~\ref{equ:word1} and Equation~\ref{equ:word2} from Equation~\ref{equ:loss2}), by removing the user and item language models for slot-value pairs (\ie removing the corresponding terms of Equation~\ref{equ:slot1} and Equation~\ref{equ:slot2} from Equation~\ref{equ:loss2}), and by removing the non-conversational term (\ie removing Equation~\ref{equ:item2} from Equation~\ref{equ:loss2}), as ConvPS$\backslash$word, ConvPS$\backslash$sv\_pair, and ConvPS$\backslash$uQv\_term, respectively. The performances of ConvPS with different components removed are shown in Figure~\ref{fig:ablation}. 
Due to the similar trends in the three categories, we illustrate the trend on the ``Cell Phones \& Accessories'' category only and omit the other two categories. For the results reported, we use GBS for the question selection. 
From what we observed in Figure~\ref{fig:ablation}, removing the user and item language models for textual descriptions (\ie ConvPS$\backslash$word), removing the user and item language models for slot-value pairs (\ie ConvPS$\backslash$sv\_pair), or removing the non-conversational term (\ie ConvPS$\backslash$uQv\_term) leads to inferior retrieval performance both before feedback (\ie the number of questions is 0) and after feedback(\ie the number of questions is from 1 to 10). This indicates that these three components are crucial. We also see that ConvPS$\backslash$sv\_pair contributes to the most significant improvement to the final performance. Removing the user and item language models for slot-value pairs leads to the highest decrease for the performance, and its performance does not increase much as the number of questions increases. This might be because removing the user and item language models for slot-value pairs greatly affects the representation learning of slot-value pairs, and thus longer conversations (more number of questions) do not obtain better performance gains. This indicates that poor representation learning introduces a risk although good representation learning of conversations leads to a benefit for the search performance~\cite{wang2021controlling,zou2020empirical,10.1145/3524110}.

Furthermore, we remove the learned representations of users, queries, slot-value pairs to perform item ranking with ConvPS during testing. We refer the ConvPS by removing the learned user representations (\ie removing the user representations from Equation~\ref{equ:24}), by removing the learned query representations (\ie removing the query representations from Equation~\ref{equ:24}), and by removing the learned slot-value pair representations (\ie removing the slot-value pair representations from Equation~\ref{equ:24}), as ConvPS$\backslash u$, ConvPS$\backslash Q$, and ConvPS$\backslash C_{qa}$, respectively. As shown in Figure~\ref{fig:ablation}, we observe that removing any of the learned representations of users, queries, or slot-value pairs leads to worse performance to retrieve items, especially when removing the learned query representations and the learned slot-value pair representations. This indicates that our learned representations are effective.

\subsection{RQ4 Parameter Sensitivity}
\label{sec:parameter}
For parameter sensitivity, we explore the effects of main parameters in our work, including (1) the impact of the number of questions, (2) the impact of embedding size, (3) the impact of batch size, and (4) the impact of explore-exploit trade-off. Due to the similar trends in the three categories, we illustrate the trend on the ``Cell Phones \& Accessories'' category only and omit the other two categories.

\subsubsection{Impact of the Number of Questions}
From Figure~\ref{fig:question}, we observe that the performance of ConvPS greatly improves as the number of questions increases. This might be because that higher number of questions gathers more information from users, leading to more close to users' actual needs. 

\begin{figure}[t]
\begin{subfigure}[t]{0.32\textwidth}
\includegraphics[width=\columnwidth]{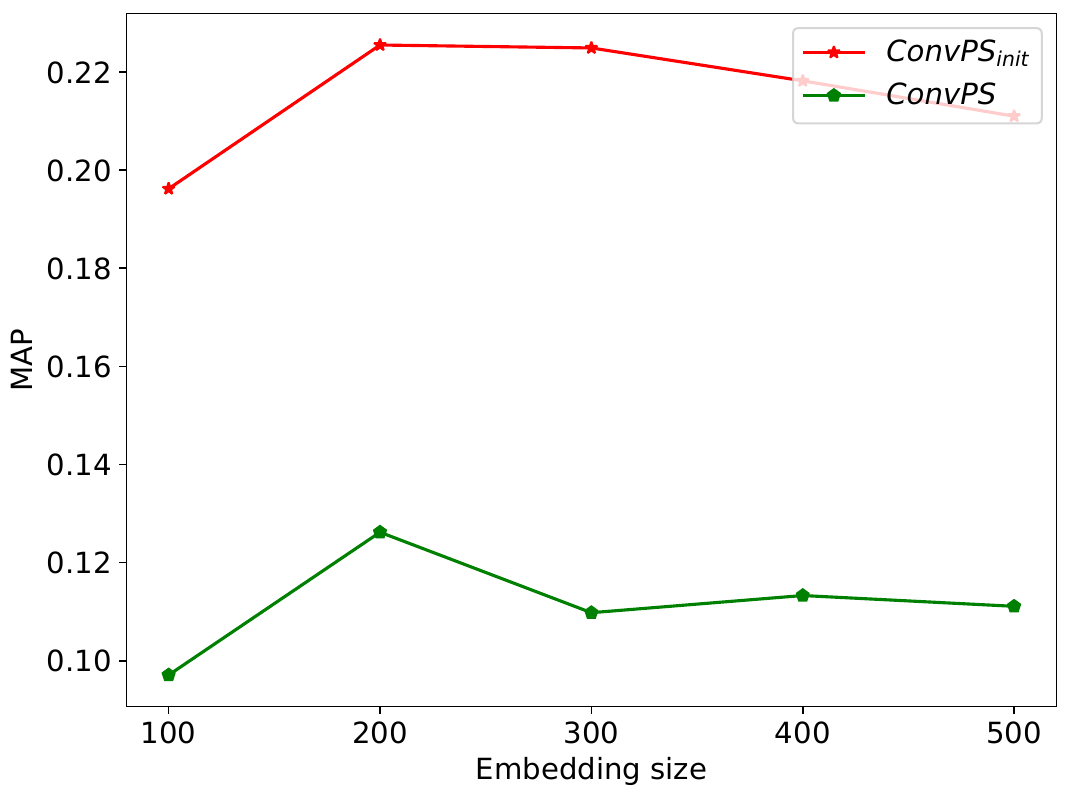}
  \caption{MAP}
  \label{fig:embedding_size1}
\end{subfigure}
\begin{subfigure}[t]{0.32\textwidth}
  \includegraphics[width=\columnwidth]{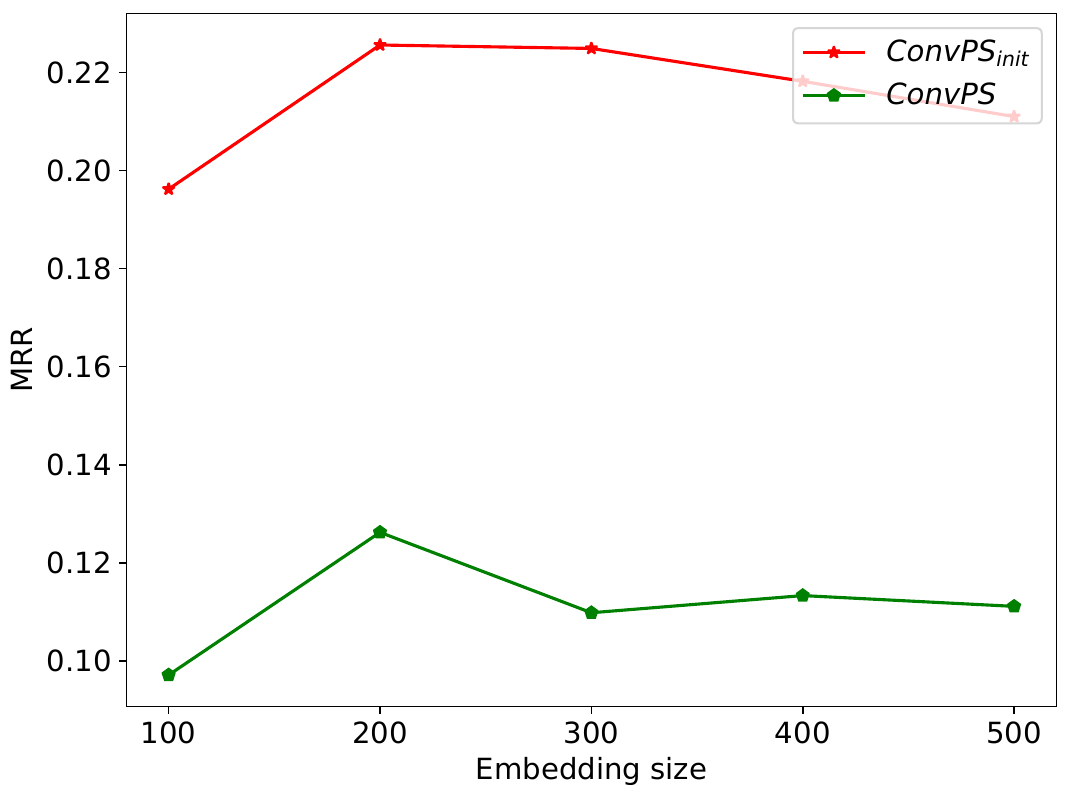}
  \caption{MRR}
  \label{fig:embedding_size2}
\end{subfigure}
\begin{subfigure}[t]{0.32\textwidth}
  \includegraphics[width=\columnwidth]{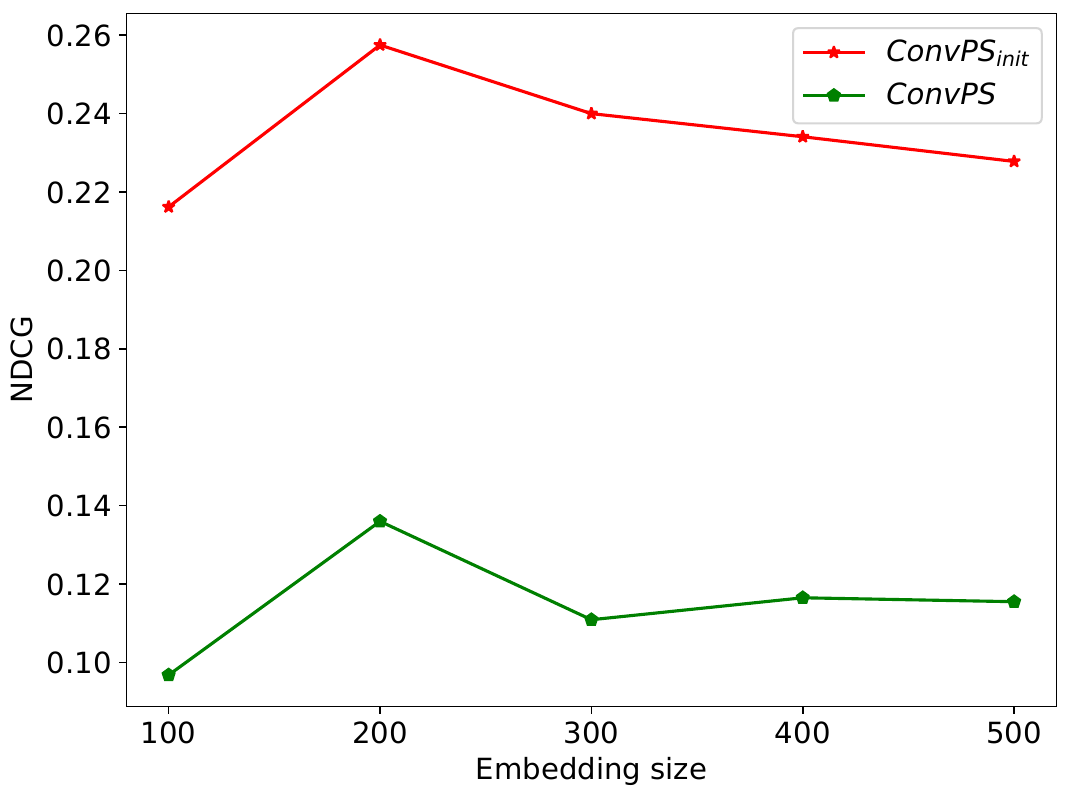}
  \caption{NDCG}
  \label{fig:embedding_size3}
\end{subfigure}
    \caption{Effect of embedding sizes.}
     \label{fig:embedding_size}
\end{figure}
\subsubsection{Impact of Embedding Size}
Figure~\ref{fig:embedding_size} shows the sensitivity of our ConvPS model in terms of embedding size. For the results reported, we use GBS for the question selection with 5 questions asked. 
As we observed, with the embedding size increasing, the performance of our ConvPS model, measured by the three metrics, first increases and then decreases. The ConvPS model achieves the best performance when the embedding size is equal to 200. Moreover, the performance gains obtained from the feedback process for our model (ConvPS vs. $\text{ConvPS}_{init}$) are stable with respect to the changes of embedding sizes.

\begin{figure}[t]
\begin{subfigure}[t]{0.32\textwidth}
\includegraphics[width=\columnwidth]{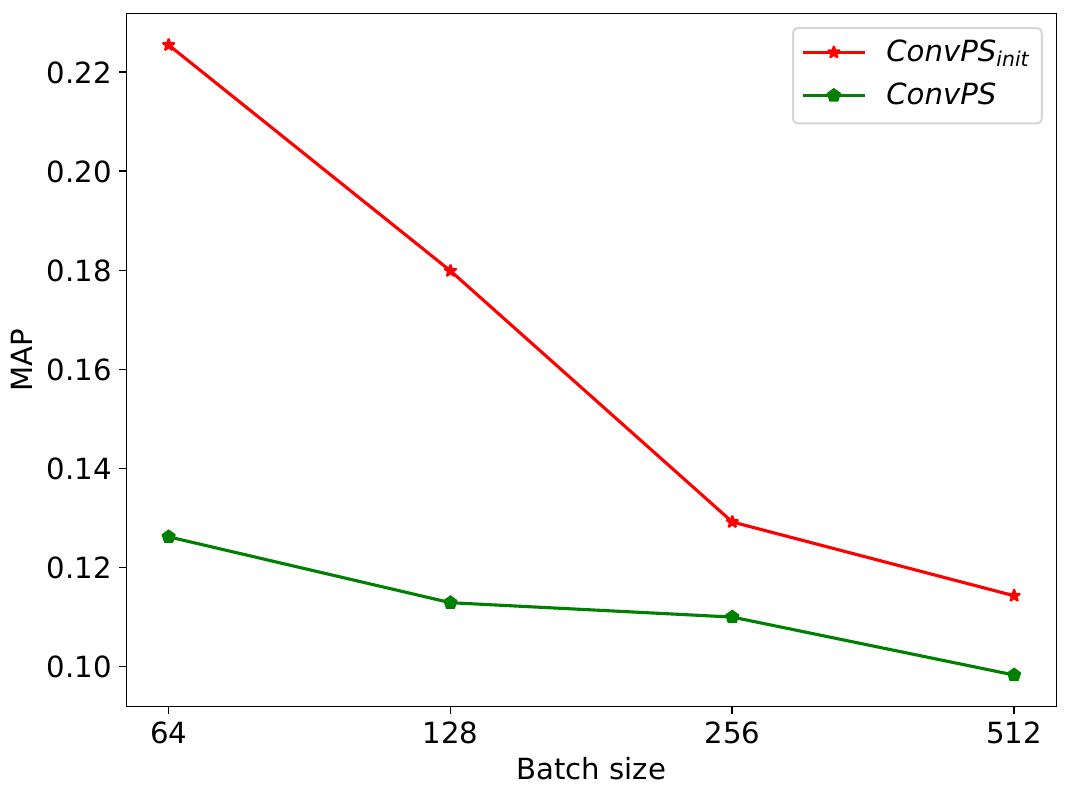}
  \caption{MAP}
  \label{fig:Batch_size1}
\end{subfigure}
\begin{subfigure}[t]{0.32\textwidth}
  \includegraphics[width=\columnwidth]{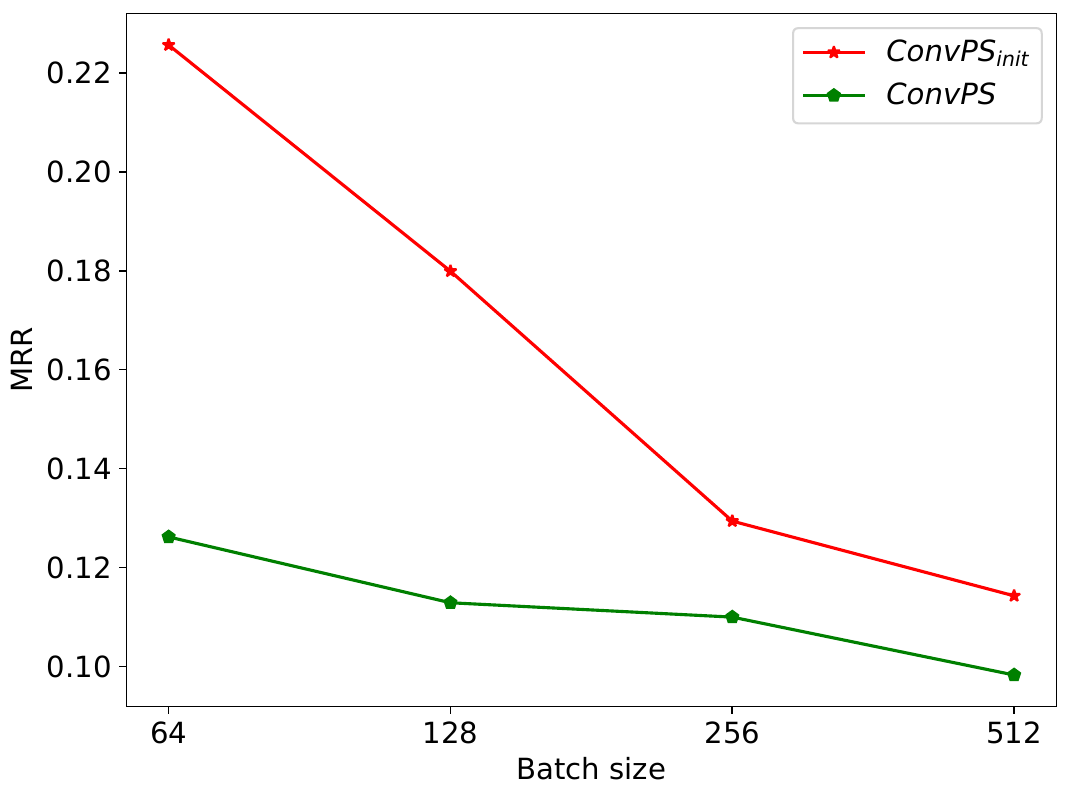}
  \caption{MRR}
  \label{fig:Batch_size2}
\end{subfigure}
\begin{subfigure}[t]{0.32\textwidth}
  \includegraphics[width=\columnwidth]{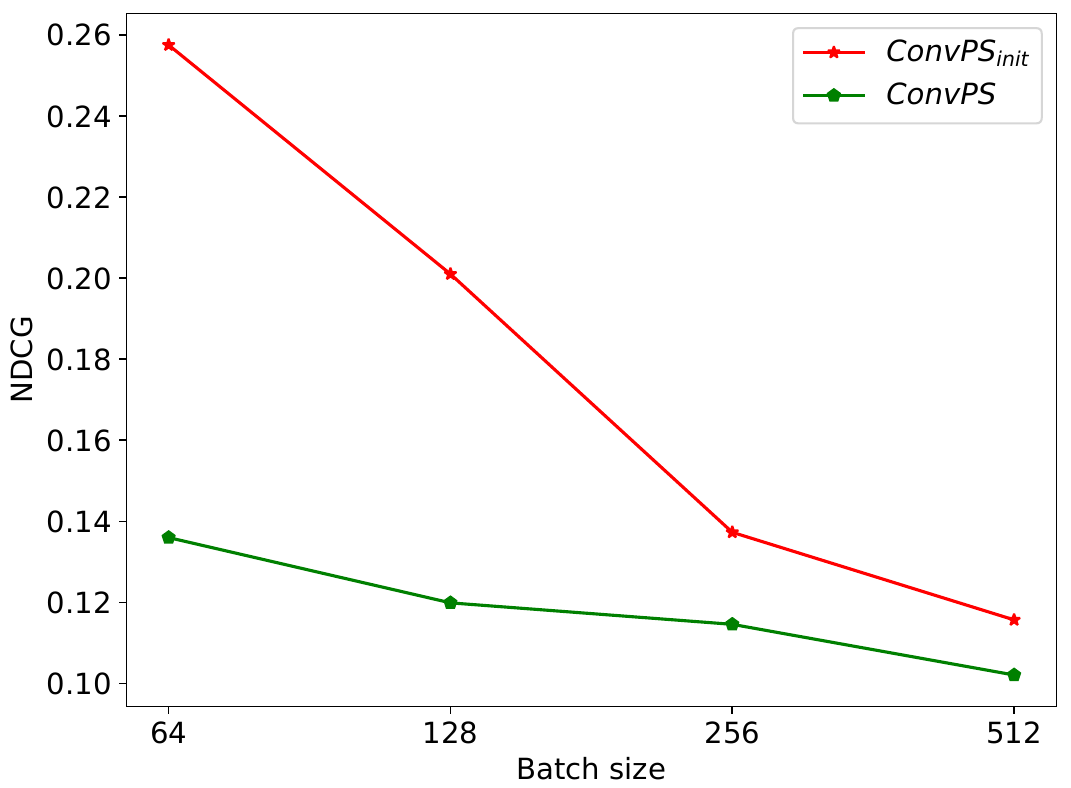}
  \caption{NDCG}
  \label{fig:Batch_size3}
\end{subfigure}
    \caption{Effect of batch sizes.}
     \label{fig:Batch_size}
\end{figure}
\subsubsection{Impact of Batch Size}
Figure~\ref{fig:Batch_size} shows the sensitivity of our ConvPS model in terms of batch size. Again, For the results reported, we use GBS for the question selection with 5 questions asked.  
We observe a decrease in the model performance on the three metrics after increasing the batch size from 64 to 512. The ConvPS model after 5 questions asked always leads to higher performance compared with $\text{ConvPS}_{init}$, and the performance gap between them becomes smaller in the case of a bigger batch size. 

\begin{figure}[t]
\begin{subfigure}[t]{0.32\textwidth}
\includegraphics[width=\columnwidth]{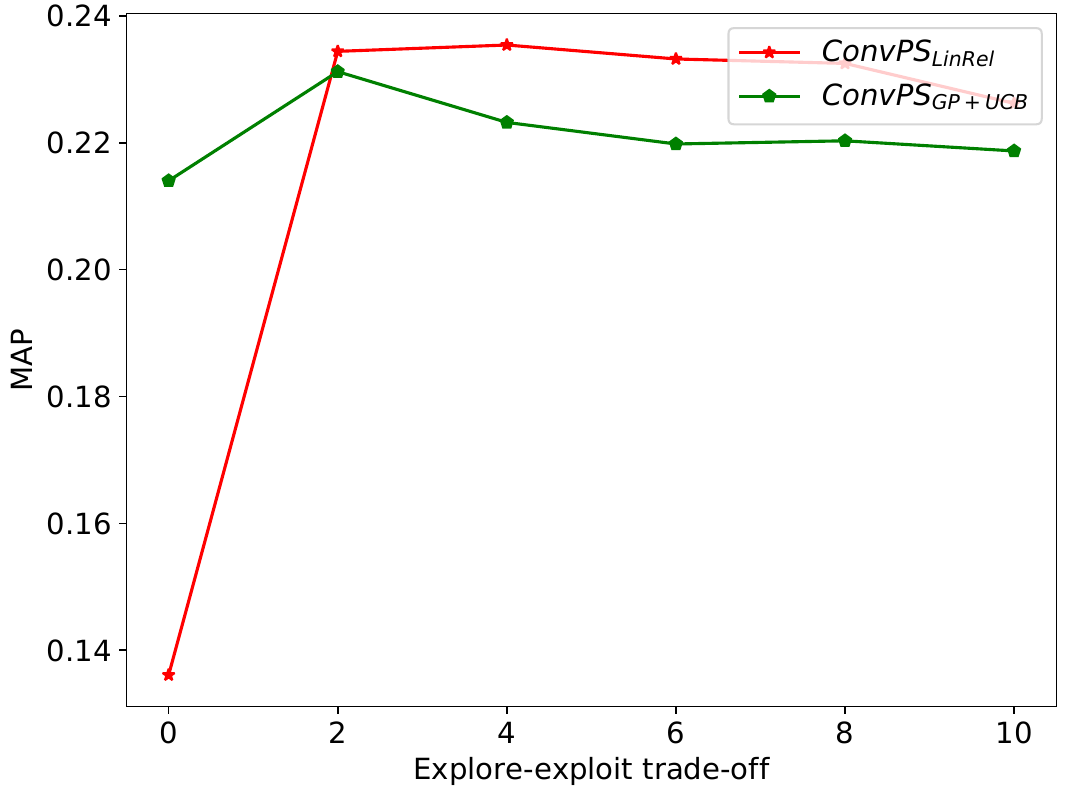}
  \caption{MAP}
  \label{fig:EE1}
\end{subfigure}
\begin{subfigure}[t]{0.32\textwidth}
  \includegraphics[width=\columnwidth]{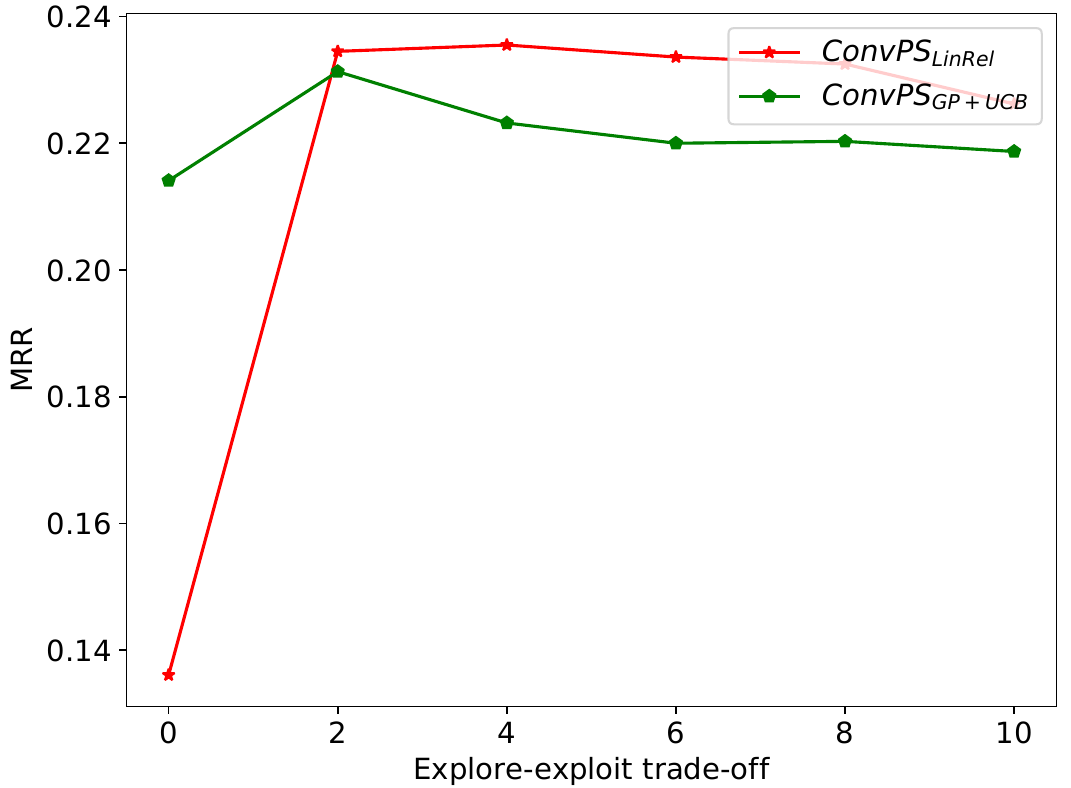}
  \caption{MRR}
  \label{fig:EE2}
\end{subfigure}
\begin{subfigure}[t]{0.32\textwidth}
  \includegraphics[width=\columnwidth]{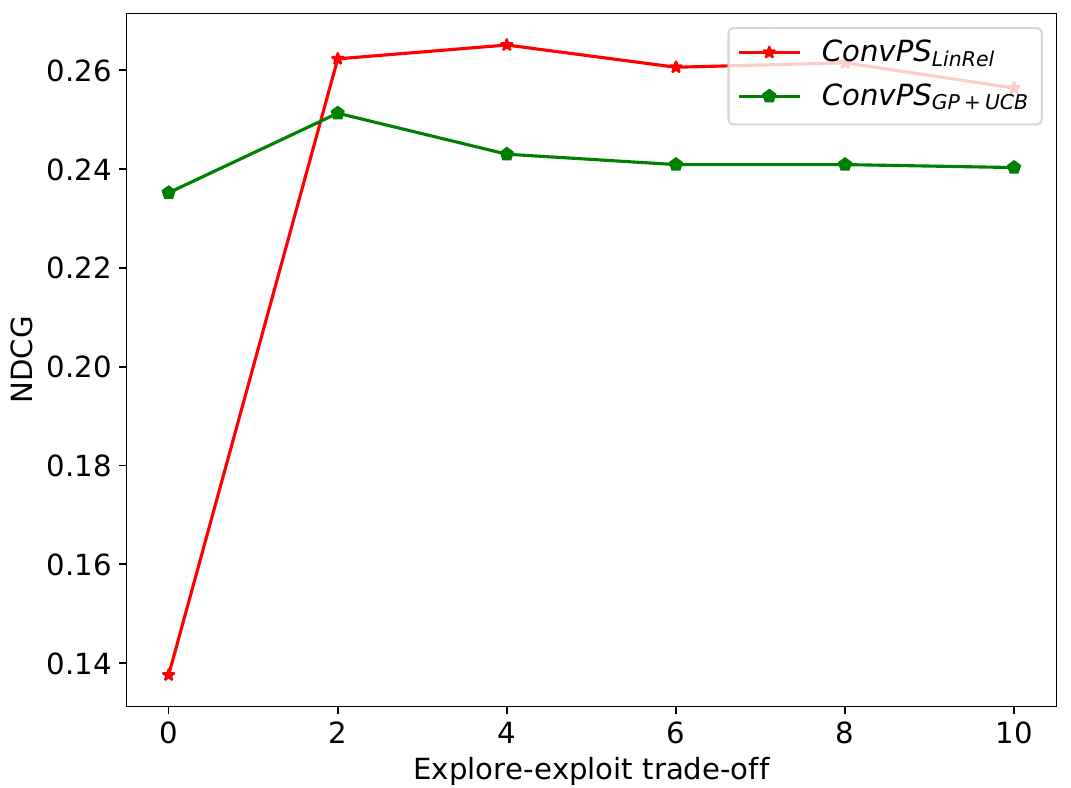}
  \caption{NDCG}
  \label{fig:EE3}
\end{subfigure}
    \caption{Effect of explore-exploit trade-off.}
     \label{fig:EE}
\end{figure}
\subsubsection{Impact of Explore-exploit Trade-off}
We explore the effect of explore-exploit trade-off in $\text{ConvPS}_{LinRel}$ and $\text{ConvPS}_{GP+UCB}$, as shown in Figure ~\ref{fig:EE}. We observe that the performance first increases and then decreases for both $\text{ConvPS}_{LinRel}$ and $\text{ConvPS}_{GP+UCB}$. The optimal explore-exploit trade-offs are 4 and 2 for $\text{ConvPS}_{LinRel}$ and $\text{ConvPS}_{GP+UCB}$, respectively. When the explore-exploit trade-off is set to zero, which means no exploration, both models achieve the lowest performance. This might be because bandit strategies like LinRel and UCB tend to ask similar questions with previous ones when there is no exploration, preventing truly effective learning from conversations, which results in a lower performance gain. This highlights the importance of exploration for learning to ask questions.  Compared with $\text{ConvPS}_{GP+UCB}$, $\text{ConvPS}_{LinRel}$ is more sensitive in the case of no exploration (\ie the explore-exploit trade-off is set to zero) while is less sensitive with the explore-exploit trade-off parameter when exploration is included (\ie the explore-exploit trade-off is larger than zero). 

\subsection{Online User Study}
By the online user study, we first examine whether real users are willing to answer the system-generated questions and how well they can answer these questions. We collected 552 conversations made between our system and 26 crowd workers\footnote{Their demographic data varied as follows: gender: 20 females, and 6 males; age: 7 participants were 18--23 years old, 3 were 24--27, 7 were 28--35, and 9 were older
than 35; English language proficiency: 20 native and 6 proficient.} on 59 target items. 
From the collected data, we see that users answered an average number of 9 questions and a median number of 10 questions per target item during the interaction with the system. In addition, in the exit questionnaire, the number of questions that users indicated they are willing to answer is 6 on average. Most of the users (64.3\%) declare that they are willing to answer at least 5 questions for locating their target products. 
For how well users can answer the system-generated questions, we observe that most (71.4\%) of the users indicate that the system’s questions were easy to answer in the exit questionnaire. In most cases, users provided the correct answers to the system’s questions (77\%). They were not sure about their answers 11\% of the time, and they provide the wrong answers (\ie their answers disagreed with the product information) 12\% of the time. 

Secondly, we evaluate the system performance, measured by the three evaluation metrics, using the collected user study data. The performance with 5 questions asked, as well as the performance when the user stopped answering questions are shown in Table~\ref{table:study}. From the results, we observe that the performance is in agreement with the offline evaluation performance of $\text{ConvPS}_{LinRel}$ in Table~\ref{table:compare}.

\begin{table}
% \captionsetup{font={small},aboveskip=3pt}
  \caption{System effectiveness on user study.}
  \centering
  \small
  \begin{tabular}{c|ccc}
    \toprule
    \# of questions & MAP & MRR & NDCG\\
    \hline
    5& 0.249 & 0.250 & 0.283\\
    stopping & 0.375 & 0.376 & 0.411\\
    \bottomrule
  \end{tabular}
  \label{table:study}
\end{table}

\section{Conclusions and Discussions}
\label{sec:conc}
In this paper, we propose a new conversational product search model, ConvPS. The ConvPS model firstly learns the representations of users, queries, items, and conversations (in terms of slot-value pairs) via a unified generative framework, and then utilizes the trained representations to retrieve the target items in the latent semantic space. Four learning to ask strategies are also incorporated into the ConvPS model to select a sequence of effective questions to ask to the user. Experimental results on the Amazon product datasets demonstrate that the conversation in product search is beneficial and there is a significant improvement of the ConvPS model compared with existing baselines. 

One limitation of this work is that we used the simulated data like previous work~\cite{bi2019conversational,zhang2018towards,zou2019learning}, because of the difficulty of obtaining such data in a real scenario. However, as early exploration in conversational product search, we demonstrate that conversational product search by incorporating user feedback on slot-value pairs via representation learning is effective and it is a promising research direction.

In this work, we construct questions by using templates on the basis of slot-value pairs. The template-based approach is simple yet effective~\cite{zamani2020generating}. However, one can also use other natural language question generation techniques to construct questions, \eg generating diverse and specific clarifying questions with keywords based on product description~\cite{zhang2021diverse}. We ask questions over slots and collect user feedback over corresponding values, as in past work in the field~\cite{zhang2018towards,sun2018conversational}. We make the assumption that the user’s feedback is positive or negative, which may not always be the case, \eg in the case the user is not certain about the answer to a question; hence, it is worth extending the system incorporating user responses other than slot-value pairs and free-form written feedback in the future work. 
Also, users may provide some custom values (along with/without the system provided values) which are out of the training vocabulary leading to an embedding mismatch. In this work we regard them as invalid questions and ignore their answers. One can also use a unified embedding or pre-trained embedding (\eg from BERT~\cite{devlin2018bert}) or use nearby embeddings to generalize the embedding of the mismatched value, which we leave for future work. 
Moreover, in this work we do not focus on natural language understanding, as done in~\citet{zhang2018towards,bi2019conversational}. We leave the natural language understanding techniques to translate user response to slot-value pairs as future work. Although slot-value pairs are effective and typically used in conversational systems (\eg slot-value pairs are usually used for dialogue state tracking in task-oriented dialogue systems)~\cite{sun2018conversational,zhong2018global}, the natural language understanding from user response to slot-value pairs may not be 100\% accurate. It is thus worth extending the system by incorporating and modeling uncertainty and noise~\cite{zou2020towardsa}.

The question selection strategies are independent (or indirectly correlated) from the trained embeddings, to maintain wide adaptability. However, one can also incorporate the trained embeddings to the objective functions of question selection strategies, so that taking the trained embeddings into account when selecting questions to ask, or simply using the trained embeddings to filter out those questions with similar embeddings based on the assumption that asking another question with similar embedding is not able to add new information to locate the target items. 

A lot of research has been done on respecting ethical implications and preserving privacy. Since our conversational product search model is personalized and based on word/slot-value embeddings, it may introduce personalization bias and embedding bias~\cite{gerritse2020bias}, leading to the system favoring some items over others. It is thus worth incorporating debiasing techniques \eg debiasing embeddings~\cite{bolukbasi2016man}, controlling personalization~\cite{ai2019zero}, and diversifying the response~\cite{gerritse2020bias} to mitigate this issue. Also, when collecting user data by asking CQs, the ethical and privacy violations may occur. A typical solution is to anonymize the data and try to hide the identity of users~\cite{carpineto2013semantic}, and encourage users to be careful about providing answers for ethical-sensitive and privacy-sensitive CQs. A privacy preserving model can also be trained when adding noise intentionally to the supervision signal to protect user privacy~\cite{duchi2014privacy}.

\section*{Acknowledgement}
\label{sec:ackn} 
We gratefully appreciate the associate editor and three anonymous reviewers for their valuable and
detailed comments that helped to improve the quality of this article. This work was partly done when the first author was a visiting Ph.D. student at the Information Retrieval and Knowledge Management Research Lab, York University, Canada. 
This research was supported by the NWO Smart Culture - Big Data / Digital Humanities (314-99-301), the NWO Innovational Research Incentives Scheme Vidi (016.Vidi.189.039), 
the H2020-EU.3.4. - SOCIETAL CHALLENGES - Smart, Green And Integrated Transport (814961), 
the Natural Science Foundation of China (62402093, 61902219), 
and 
the Google Faculty Research Awards program. This research was also supported by the research grant (RGPIN-2020-07157) from Natural Sciences and Engineering Research Council (NSERC) and York Research Chairs (YRC) program. 
All contents represent the opinion of the authors, which is not necessarily shared or endorsed by their respective employers and/or sponsors. We also acknowledge Compute Canada for providing us with the computing resources to conduct experiments. 
%\clearpage
%\newpage

\bibliographystyle{ACM-Reference-Format}
\bibliography{bibfile} 

\end{document}